\documentclass{article} 
\usepackage{iclr2024_conference,times}

\usepackage{amsmath,amsfonts,bm}

\def\eqref#1{equation~\ref{#1}}

\def\1{\bm{1}}

\DeclareMathAlphabet{\mathsfit}{\encodingdefault}{\sfdefault}{m}{sl}
\SetMathAlphabet{\mathsfit}{bold}{\encodingdefault}{\sfdefault}{bx}{n}

\usepackage[colorlinks=true,linkcolor={brown}, citecolor={brown}]{hyperref}%
\usepackage{url}

\usepackage{algorithm}
\usepackage{algorithmic}
\usepackage[utf8]{inputenc} 
\usepackage[T1]{fontenc}    
\usepackage{url}            
\usepackage{booktabs}       
\usepackage{amsfonts}       
\usepackage{nicefrac}       
\usepackage{microtype}      
\usepackage{bm}
\usepackage{enumitem}
\usepackage{graphicx}
\usepackage{amsmath}
\usepackage{subfigure}
\usepackage{amssymb}
\usepackage{caption}

\usepackage{comment}
\usepackage{xcolor}         
\usepackage{centernot}

\usepackage{array}
\usepackage{multirow}
\usepackage{wrapfig,lipsum,booktabs}
\usepackage{makecell}
\usepackage{pifont}

\newcommand{\dummy}[1]{{\textcolor{black}{#1}}}
\newcommand{\hl}[1]{{\textcolor{orange}{#1}}}

\usepackage{tabularx}

\usepackage{newfloat}
\usepackage{listings}
\newcommand{\tabitem}{~~\llap{\textbullet}~~}

\definecolor{eclipseStrings}{RGB}{42,0.0,255}
\definecolor{eclipseKeywords}{RGB}{127,0,85}
\colorlet{numb}{magenta!60!black}

\lstdefinelanguage{json}{
    basicstyle=\normalfont\ttfamily,
    commentstyle=\color{eclipseStrings}, 
    stringstyle=\color{eclipseKeywords}, 
    numbers=left,
    numberstyle=\scriptsize,
    stepnumber=1,
    numbersep=8pt,
    showstringspaces=false,
    breaklines=true,
    frame=lines,
    backgroundcolor=\color{white}, 
    string=[s]{"}{"},
    comment=[l]{:\ "},
    morecomment=[l]{:"},
    literate=
        *{0}{{{\color{numb}0}}}{1}
         {1}{{{\color{numb}1}}}{1}
         {2}{{{\color{numb}2}}}{1}
         {3}{{{\color{numb}3}}}{1}
         {4}{{{\color{numb}4}}}{1}
         {5}{{{\color{numb}5}}}{1}
         {6}{{{\color{numb}6}}}{1}
         {7}{{{\color{numb}7}}}{1}
         {8}{{{\color{numb}8}}}{1}
         {9}{{{\color{numb}9}}}{1}
}

\title{DCR-Consistency: Divide-Conquer-Reasoning for Consistency Evaluation and Improvement of Large Language Models}

\newcommand{\methodName}{\texttt{DCR} }
\newcommand{\methodNamenew}{\texttt{DCR}}
\newcommand{\divideconquer}{\texttt{DCE}}
\newcommand{\metric}{\texttt{AMC}}
\newcommand{\reasoning}{\texttt{RAI}}

\author{Wendi Cui$^1$, \quad  Jiaxin Zhang$^{1,2}$\thanks{Corresponding Author. Our code is available at \url{https://github.com/intuit-ai-research/DCR-consistency}.}, \quad Zhuohang Li$^3$, \quad Damien Lopez$^1$, \quad Kamalika Das$^{1,2}$, \\ {\bf Bradley Malin$^{3,4}$, \quad Sricharan Kumar$^{1,2}$}
\\ $^1$Intuit \quad  $^2$Intuit AI Research \quad $^3$Vanderbilt University \quad $^4$Vanderbilt University Medical Center \\
\texttt{\{wendi\_cui, jiaxin\_zhang\}@intuit.com}, \quad 
\texttt{zhuohang.li@vanderbilt.edu}
}

\iclrfinalcopy 
\begin{document}

\maketitle

\begin{abstract}
Evaluating the quality and variability of text generated by Large Language Models (LLMs) poses a significant, yet unresolved research challenge. Traditional evaluation methods, such as ROUGE and BERTScore, which measure token similarity, often fail to capture the holistic semantic equivalence. This results in a low correlation with human judgments and intuition, which is especially problematic in high-stakes applications like healthcare and finance where reliability, safety, and robust decision-making are highly critical. {\color{black}This work proposes \methodNamenew, an automated framework for evaluating and improving the consistency of LLM-generated texts using a divide-conquer-reasoning approach.} Unlike existing LLM-based evaluators that operate at the paragraph level, our method employs a divide-and-conquer evaluator (\divideconquer)
that breaks down the paragraph-to-paragraph comparison between two generated responses into individual sentence-to-paragraph comparisons, each evaluated based on predefined criteria. To facilitate this approach, we introduce an automatic metric converter (\metric) that translates the output from \divideconquer~into an interpretable numeric score. Beyond the consistency evaluation, we further present a reason-assisted improver (\reasoning) that leverages the analytical reasons with explanations identified by \divideconquer~to generate new responses aimed at reducing these inconsistencies. Through comprehensive and systematic empirical analysis, we show that our approach outperforms state-of-the-art methods by a large margin (e.g., +{\bf 19.3}\% and +{\bf 24.3}\% on the SummEval dataset) in evaluating the consistency of LLM generation across multiple benchmarks in semantic, factual, and summarization consistency tasks. Our approach also substantially reduces nearly 90\% output inconsistencies, showing promise for effective hallucination mitigation. 
\end{abstract}

\section{Introduction}
Large language models (LLMs) such as GPT-4 and PaLM 2 \citep{yang2023harnessing,bubeck2023sparks} have demonstrated impressive performance on a variety of natural language generation (NLG) tasks, including summarization \citep{tam2022evaluating}, open-book question-answering (QA) \citep{kamalloo2023evaluating}, and retrieval-augmented generation (RAG) \citep{lewis2020retrieval, liu2023reta}. The evaluation of generated response quality often involves the assessment of the semantic equivalence between two pieces of text, e.g., between the generated response and the original text in summarization tasks or between two candidate responses in open-book QA tasks.
However, conventional evaluation methods, such as BARTScore \citep{yuan2021bartscore} and BERTScore \citep{zhang2019bertscore}, which rely on {\em token-level} comparison, are inadequate for accurately and reliably measuring the quality of generated content, particularly in complex scenarios with long paragraphs \citep{liu2023gpteval, hanna2021fine}. To address this issue, LLM-based evaluators such as G-Eval \citep{liu2023gpteval} and GPTScore \citep{fu2023gptscore} have proposed a new framework that evaluates texts via {\em paragraph-level} comparison. While these evaluators show promise for certain tasks, their scores often fail to achieve high concordance with human judgments of semantic equivalence. Furthermore, as only numeric scores are provided with no explanation, it can be challenging for humans to trust or reason about these scores, particularly when using LLMs that are known to hallucinate \citep{li2023halueval, ji2023survey, rawte2023survey}.

Assessing the consistency of LLMs is more broadly connected to AI safety and has become a critical step in improving the reliability of these systems by preventing the generation of misinformation and harmful content. \citet{wang2022self} demonstrates that {\em consistency checking} can significantly enhance the chain of thought reasoning in LLMs. Similarly, \citet{kuhn2023semantic} leverages semantic consistency for uncertainty estimation in NLG. Recent studies employ consistency checking to detect hallucinations based on pre-trained LLMs \citep{manakul2023selfcheckgpt,zhang2023sac} and instruction-tuned LLMs \citep{mundler2023self}. Although these methods exhibit promising results on several specific tasks, including mathematical reasoning and factual assessment, the potential failures \citep{chen2023two} of self-consistency are often overlooked. This is essentially due to a lack of a generic, automatic, and reliable strategy that assesses the consistency of two responses, let alone remediating such inconsistency after identifying them.
\begin{figure}[h!]
\centering
    \includegraphics[width=0.99\textwidth]{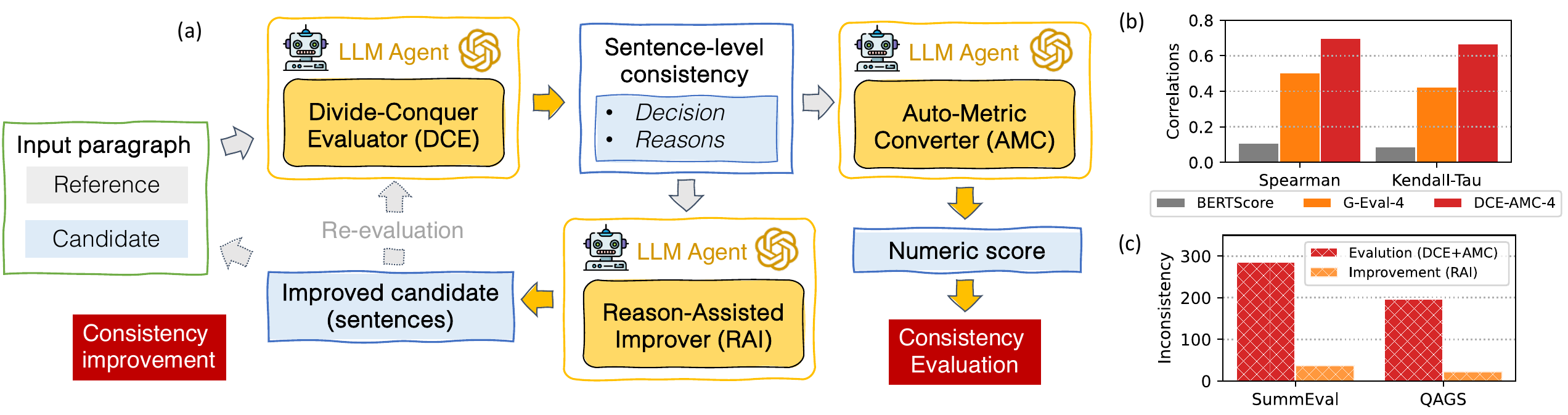}
    \caption{(a) Overview of the proposed \methodName framework. The first two components (\divideconquer-\metric) aim at providing a better strategy for evaluating and quantifying semantic consistency to best match human judgments. Building on this, a third component \reasoning~further utilizes analytical reasoning to iteratively improve the consistency of LLM-generated content w.r.t. the reference by minimizing hallucinations. (b) The combination of \divideconquer~and \metric~(\texttt{DCE-AMC}-4) significantly outperforms the baseline methods in terms of correlations with human ratings. (c) \reasoning~substantially reduces output inconsistencies by $\sim90\%$ through a single improvement iteration on SummEval and QAGS benchmarks.}
    \label{fig:overview0}
\end{figure}
\vspace{-0.3cm}

In this paper, we introduce a novel framework, called \texttt{Divide-Conquer-Reasoning} (abbreviated as \methodNamenew~hereafter), for developing an automatic and reliable consistency evaluation method. Our approach capitalizes on the intuition that human evaluators typically assess consistency by comparing the semantic meaning of the generated text to the reference text sentence by sentence and then combining the analysis to make a holistic judgment of the complete concept. Unlike existing metrics that rely on either token-level or paragraph-level checks, 
our approach breaks down the paragraph-to-paragraph comparison of the two pieces of text into a series of sentence-to-paragraph comparisons. This approach avoids confusing LLM by either providing too much information at once or zooming in too narrowly. Additionally, our approach does not rely on LLMs to directly output verbal scores, which have been shown to be prone to hallucination. We note that our \metric~ method is a reference-free method, which does not rely on a golden reference written by the human expert. For example in a summary task, our DCR method does not need a sample summary and can compare directly between the target summary and the original paragraphs.


The \methodNamenew~framework is composed of three components, each executed by an LLM agent, as shown in Fig.~\ref{fig:overview0}. Given the reference and candidate, the Divide-Conquer Evaluator (\divideconquer) realizes the notion of divide-conquer to determine whether the candidate is semantically equivalent to the reference at a sentence level. \divideconquer~automatically partitions the candidate paragraph into sentences ({\em divide}), evaluates each sentence against the reference paragraph based on pre-defined semantic-level consistency criteria ({\em conquer}), and generates a list of reasons that explain why each sentence is or is not consistent with the reference paragraph. Next, the Auto-Metric Converter (\metric) which builds upon \divideconquer, converts the reasons (with explanations) into a numeric score 
that is more aligned with human intuition and provides better interpretation. The numeric score can be used to evaluate consistency in various tasks, such as summarization, factual assessment, and hallucination detection.

Our \methodNamenew~framework not only evaluates consistency but also enhances it through the Reason-Assisted Improver (\reasoning), a third LLM agent that utilizes the outputs of \divideconquer~to generate new candidate sentences. By incorporating the explanations provided by \divideconquer~with the original context, \reasoning~produces sentences that mitigate inconsistencies (hallucinations). This improvement process can be iteratively applied by utilizing the re-evaluation produced by \divideconquer~to ultimately achieve a candidate response that is fully aligned with the reference text.

We conducted an evaluation of our approach on three different NLG tasks, including semantic, summarization, and factual consistency evaluations. Our results demonstrate that \methodName significantly outperforms all existing baseline methods as a consistency evaluator, with improvements of up to 19.3\% and 24.3\% compared to G-Eval on the SummEval dataset. Additionally, our approach achieved high correlations with human judgment on all three benchmarks. Notably, we observed highly promising results in consistency improvement rate (from 86.71\% to 91.11\%) at a substantially lower effort and cost due to its multi-thread parallel implementation.

\section{Preliminaries}

\paragraph{Limitation of Existing Methods.}
The conventional metrics, such as BERTscore and BARTscore, rely on a {\em token-level} comparison using n-gram or contextual embedding to calculate cosine similarity. However, this approach fails to capture the overall semantic meaning as it directly aggregates token-level similarities. To address this issue, leveraging the power of LLMs for self-evaluation has been proposed. G-Eval \citep{liu2023gpteval} and GPT-Eval \citep{jiang2023llm} evaluate consistency at a paragraph level by prompting LLMs to compare two candidates as a whole. However, these approaches have a major drawback as the generated verbal scores by LLMs are {\em prone to hallucinations}, resulting in abnormally higher ratings for LLM-generated content that diverge from human judgment \citep{liu2023gpteval}. Such methods also generate no actionable insight to justify the score or mitigate inconsistencies after identifying them.

\paragraph{Problem Formulation.}
Given a user query $\mathcal{Q}$ and LLM model $\mathcal{M}$, let $\mathcal{C}$ refer to the candidate response drawn from $\mathcal{C} = \mathcal{M}(\mathcal{Q})$. LLM-generated responses are commonly evaluated using some reference texts, denoted by $\mathcal{R}$, for instance, human writing samples for generation tasks and original content for summarization tasks. The objective of consistency evaluation is to build a function $f$ that quantitatively measures
the {\em semantic equivalence} $\mathcal{S}$ between the generated candidates $\mathcal{C}$ and reference $\mathcal{R}$ as $\mathcal{S} = f(\mathcal{R},\mathcal{C} | \mathcal{Q}, \mathcal{M})$ where $\mathcal{S}$ could be binary decision, such as ``Yes'' or ``No'', ``Consistent'' or ``Not Consistent'', or numeric score, e.g., [-1, +1]. However, it is worth noting that our evaluation can be generally used to check consistency between two candidates where both are generated by LLMs. In that scenario, we only need to assume one candidate as the reference for self-check consistency. {\color{black} Note that our proposed method employs the same setting as G-Eval \citep{liu2023gpteval}, which does not rely on a golden reference written by the human expert or ground truth labels. Our objective is to offer a generic way to assess the consistency between two text sequences. For instance, in the summarization task, one text sequence may be the original document, while the other might be the generated text summarization by LLMs. To distinguish between the two text sequences, we referred to one of the text sequences as “reference” and the other as “candidate”. }

\begin{figure*}[t]
\centering
    \includegraphics[width=0.99\linewidth]{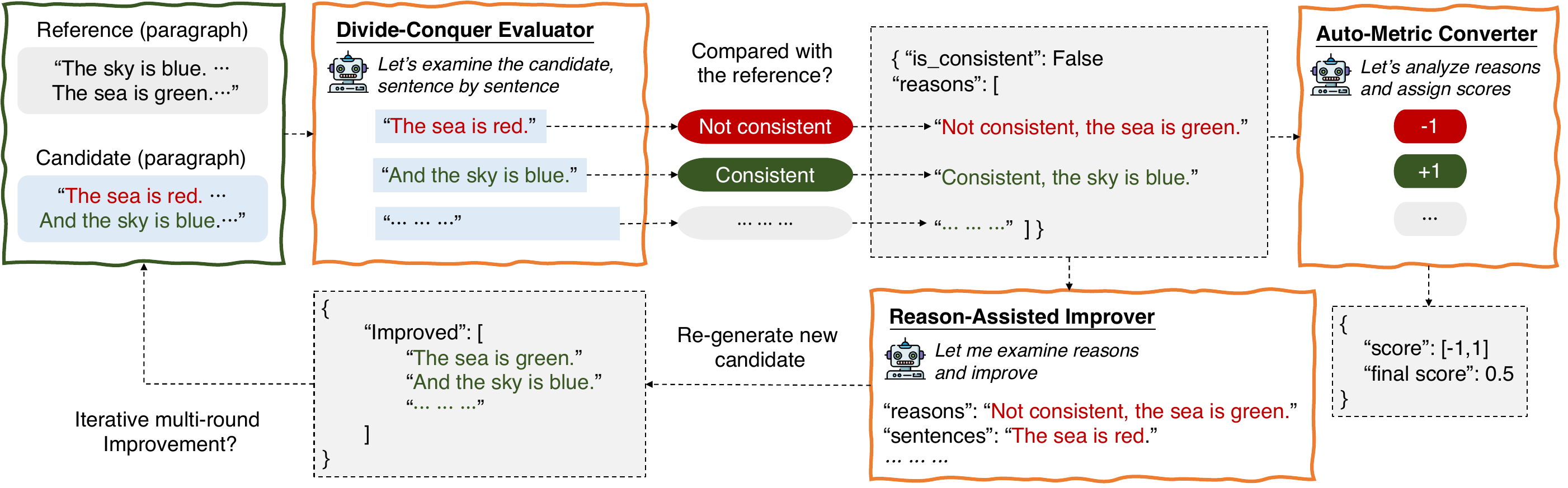}
    \vspace{-2mm}
    \caption{An example of evaluating and improving the consistency of generated text via \methodNamenew.}
    \label{fig:div_example}
    \vspace{-3mm}
\end{figure*}

\section{Divide-Conquer-Reasoning}
To overcome the aforementioned limitations, we propose to evaluate and improve the consistency of LLM output via a Divide-Conquer-Reasoning approach, referred to as \methodNamenew. The approach comprises three key components, as illustrated in Fig.~\ref{fig:overview0}: (1) \divideconquer, which disassembles the candidate paragraph and scrutinizes semantic inconsistencies sentence-by-sentence, (2) \metric, which converts sentence-level inconsistency/consistency reasons into numeric scores for quantitative interpretation, and (3) \reasoning, which conducts analytical reasoning to improve consistency through candidate regeneration. Essentially, our approach involves a combination of sentence-level analysis, semantic consistency checking, and causal analysis, making it an ideal evaluation metric for a diverse range of NLG tasks that require comparison to reference texts, such as summarization, open-book question-answering (QA), and retrieval-augmented generation.

Moreover, \methodName not only evaluates but also improves the consistency of generated text through analysis and reasoning, which aligns with human intuition. Fig.\ref{fig:div_example} provides an example of how \methodName~can evaluate and enhance the consistency of candidate text. In the following sections, we will discuss each component in detail.

\subsection{Divide-Conquer Evaluator (\divideconquer)} 
The Divide-Conquer Evaluator (\divideconquer) is an LLM Agent designed to perform semantic consistency checks between the reference and the candidate using a sentence-by-sentence strategy. This agent accepts a reference paragraph and a candidate paragraph as inputs, and employs a divide-conquer strategy to break down the entire paragraph into multiple individual sentences ({\em divide}) and then assess each sentence against the reference ({\em conquer}). More specifically, given the input reference $\mathcal{R} = \langle s_1^r, ...,s_l^r \rangle$ and candidate $\mathcal{C}= \langle s_1^c, ...,s_k^c \rangle$, we build a \divideconquer~agent $\mathcal{L}_{\textup{\divideconquer}}$ using the LLM model $\mathcal{M}$ (e.g., GPT-3.5/4) with an instructed prompt $\mathcal{P}_{\textup{\divideconquer}}$ as:
\begin{equation}
    \{ \gamma_1, \gamma_2,...,\gamma_k \}= \mathcal{L}_{\textup{\divideconquer}}(\langle s_1^c, s_2^c,...,s_k^c \rangle, \mathcal{R} ~|~ \mathcal{M}, \mathcal{P}_{\textup{\divideconquer}}).
    \label{eq:dce}
\end{equation}
Eq.\ref{eq:dce} generates {\em reasons}, denoted as $\Gamma = \{ \gamma_1, \gamma_2,...,\gamma_k \} $, which is a list of reasons explaining why each sentence $s_i^c (i=1,2,...,k)$ is or is not consistent against the {\em entire} reference paragraph $\mathcal{R}$. It's important to note that the reasons $\gamma_i$ might comprise a short paragraph containing multiple explanation sentences. We can tailor instruction prompts by defining task-specific criteria to accommodate different tasks. Table \ref{tab:prompt} provides an example of a prompt example with pre-defined criteria for the summarization consistency task.

\begin{table}[h!]
\centering
\caption{Summarization Consistency Divide-Conquer Evaluator Prompt}
\vspace{-0.3cm}
\label{tab:prompt}
\begin{tabular}{p{13.5cm}}
\toprule 
Your task is to evaluate whether the summary is consistent with the article. You will evaluate it by going through each sentence of the summary and check against the following procedures:  \\
\vspace{0.5pt}
\tabitem {\em Understands all the aspects of the sentence, and compare if each aspect exists in the article}  \\
\tabitem {\em If it does, compare if the information in this sentence is consistent with what is in the article}  \\
\tabitem {\em Compare if all the information in this sentence can be directly inferred or entailed from what is in the article. It is OK that not all information from the article exists in this summary }
\\
\bottomrule
\end{tabular}
\end{table}

{\color{black}We note that the sentence-level comparison used in \divideconquer~is not to compare each sentence in the candidate text sequence to each sentence from the reference text sequence but to compare each sentence in the candidate text sequence to the {\em entire} reference text sequence. This design reduces the number of comparison operations and ensures semantic consistency. For instance, in the summarization task, \divideconquer~is to check if the candidate summaries are consistent with the source article. This is achieved by checking if each sentence in the candidate summaries is consistent with the {\em entire} original articles, rather than a sentence-by-sentence matching. Thus, our method does not rely on any sentence-matching techniques, such that there is no issue with the varying number of sentences \citep{amplayo2022smart}.}

\subsection{Auto-Metric Converter (\metric)} 
The Auto-Metric Converter (\metric) is an LLM Agent that aims to quantitatively measure the consistency evaluation derived from the Divide-Conquer Evaluator (\divideconquer) by converting the reasons from \divideconquer~into a numeric score system. This is accomplished by introducing an LLM agent, denoted as $\mathcal{L}_{\textup{\metric}}$, which takes reasons $\langle \gamma_1, \gamma_2,...,\gamma_k \rangle$ with an instructed prompt $\mathcal{P}_{\textup{\metric}}$ as inputs:
\begin{equation}
    \{ z_1, z_2,...,z_k \}= \mathcal{L}_{\textup{\metric}}(\{ \gamma_1, \gamma_2,...,\gamma_k \} ~|~ \mathcal{M}, \mathcal{P}_{\textup{\metric}}).
    \label{eq:amc}
\end{equation}
The LLM Agent $\mathcal{L}_{\textup{\metric}}$ functions as a binary sentiment classifier that classifies the reasons $\langle \gamma_1, \gamma_2,...,\gamma_k \rangle$ to be either positive (marked by ``+1" if the sentence is consistent), or negative (marked by ``-1'' otherwise). As a result, \metric~outputs an array of scores $\{ z_1, z_2,...,z_k \}, z_i \in \{-1, +1\}$ for each sentence $\langle s_1^c, s_2^c,...,s_k^c \rangle$ in the candidate $\mathcal{C}$. We then utilize this score array to calculate a comprehensive score $\mathcal{Z}$ to evaluate how consistent the candidate (paragraph) is against the reference (paragraph): 
\begin{equation}
    \mathcal{Z} = \left(\sum_{i=1}^k z_i + \alpha \right) / (k + \beta), \quad \hat{\mathcal{Z}} = (\mathcal{Z} + 1)/2, \quad \hat{\mathcal{Z}} \in [0,1] \label{eq:amc-final}
\end{equation}
where $k$ is the length of the score array, i.e., the number of sentences in the candidate paragraph. Depending on the prompt, the {\em reasons} output by \divideconquer~may not all be on the sentence level. To ensure that the score calculated is solely generated by sentence-level {\em reasons}, we introduce $\alpha$ and $\beta$ in Eq.~\ref{eq:amc-final}, as explained in detail in Appendix \ref{sec:alpha_beta}. Finally, we rescale $\mathcal{Z}$ to obtain the final score $\hat{\mathcal{Z}}$ that is typically between 0 ({\em completely inconsistent}) and 1 ({\em completely consistent}). The closer this score $\hat{\mathcal{Z}}$ is to 0, the more inconsistent the candidate $\mathcal{C}$ is against the reference $\mathcal{R}$.  

{\color{black}The \metric~component serves as a binary sentiment classifier that classifies the reasons to be either positive (marked by “+1" if the sentence is consistent), or negative (marked by “-1” otherwise) for each sentence, then utilizes this score array to calculate a comprehensive score to evaluate consistency in a regression manner}. {\color{black}Such a numerical score calculated by AMC is more stable than the verbal score directly output by LLMs, as is verified by our experiments.
This design deliberately excludes the use of LLM in crucial steps associated with generating numerical evaluation scores. Instead, it harnesses the capabilities of LLMs for tasks in which they have demonstrated excellence, specifically in reasoning.}

\subsection{Reason-Assisted Improver (\reasoning)} 
The Reason-Assisted Improver (\reasoning) is an LLM Agent that focuses on improving the consistency of candidate sentences by reasoning through the inconsistent explanations generated by the Divide-Conquer Evaluator (\divideconquer). To achieve this goal, we propose an LLM agent $\mathcal{L}_{\textup{\reasoning}}$ to generate new candidate sentences $\langle \hat{s}_1^c, \hat{s}_2^c,...,\hat{s}_k^c \rangle$ based on the  collected reasons $\{ \gamma_1, \gamma_2,...,\gamma_k \}$ and original sentences $\langle s_1^c, s_2^c,...,s_k^c \rangle$:
\begin{equation}
   \langle \hat{s}_1^c, \hat{s}_2^c,...,\hat{s}_k^c \rangle = \mathcal{L}_{\textup{\reasoning}}(\{ \gamma_1, \gamma_2,...,\gamma_k \}, ~ \langle s_1^c, s_2^c,...,s_k^c \rangle, \mathcal{R} ~|~ \mathcal{M}, \mathcal{P}_{\textup{\reasoning}}).
    \label{eq:rai}
\end{equation}
The core task of $\mathcal{L}_{\textup{\reasoning}}$ is to rewrite the original sentence $s_i^c$ if $s_i^c$ is inconsistent with the reference $\mathcal{R}$ and return a new generated $\hat{s}_i^c$ ($\hat{s}_i^c \neq s_i^c$), otherwise retain $s_i^c$. The newly generated responses $\hat{\mathcal{C}} = \langle \hat{s}_1^c, \hat{s}_2^c,...,\hat{s}_k^c \rangle$ can be considered as the consistency-improved candidate, which can be re-evaluated by \divideconquer~to check if $\hat{\mathcal{C}}$ mitigates inconsistencies against the reference $\mathcal{R}$.  

The improved candidate $\hat{\mathcal{C}}$ in Eq.\ref{eq:rai} can be directly fed to the \divideconquer~agent in Eq.\ref{eq:dce} after the {\em first-round} \methodName, i.e., \divideconquer $\rightarrow$ \metric $\rightarrow$ \reasoning. A straightforward extension is {\em multi-round} consistency improvement, where the consistency is iteratively improved until reaching the maximum number of rounds $m$. Algorithm \ref{alg_self} illustrates the workflow of the \methodName~framework, which consists of three core components: \divideconquer, \metric, and \reasoning.

\begin{algorithm}[h!]
\small
\caption{Proposed \texttt{Divide-Conquer-Reasoning} (\methodNamenew) framework}
\label{alg_self}
\begin{algorithmic}
\STATE {\bf Requirements}: Candidate $\mathcal{C}$, Reference $\mathcal{R}$, LLM model $\mathcal{M}$, LLM agents $\mathcal{L}_{\textup{\divideconquer}}$, $\mathcal{L}_{\textup{\metric}}$, $\mathcal{L}_{\textup{\reasoning}}$ with instructed prompts $\mathcal{P}_{\textup{\divideconquer}}$,  $\mathcal{P}_{\textup{\metric}}$ and $\mathcal{P}_{\textup{\reasoning}}$, and the maximum number of rounds $m$ 
\STATE {\bf for} {rounds $r=1,...,m$}  {\bf do}
\STATE \quad Disassemble candidate $\mathcal{C}$ into sentences $\langle s_1^c, s_2^c,...,s_k^c \rangle$, evaluate sentence-level consistency against \\ \quad reference $\mathcal{R}$, and return the reasons $\{ \gamma_1, \gamma_2,...,\gamma_k \} \leftarrow \mathcal{L}_{\textup{\divideconquer}}(\langle s_1^c, s_2^c,...,s_k^c \rangle, \mathcal{R} ~|~ \mathcal{M}, \mathcal{P}_{\textup{\divideconquer}})$ in Eq.~\ref{eq:dce}
\STATE \quad Transform reasons into numeric scores $\{ z_1, z_2,...,z_k \} \leftarrow \mathcal{L}_{\textup{\metric}}(\{ \gamma_1, \gamma_2,...,\gamma_k \} ~|~ \mathcal{M}, \mathcal{P}_{\textup{\metric}})$ in Eq.~\ref{eq:amc}
\STATE \quad Calculate the final consistency evaluation score $\hat{\mathcal{Z}}$ based on $\{ z_1, z_2,...,z_k \}$ using Eq.~\ref{eq:amc-final}
\STATE \quad Generate improved candidate $\langle \hat{s}_1^c, \hat{s}_2^c,...,\hat{s}_k^c \rangle \leftarrow  \mathcal{L}_{\textup{\reasoning}}(\{ \gamma_1, \gamma_2,...,\gamma_k \}, ~ \langle s_1^c, s_2^c,...,s_k^c \rangle, \mathcal{R} ~|~ \mathcal{M}, \mathcal{P}_{\textup{\reasoning}})$ 
\STATE \quad Update the candidate $\langle s_1^c, s_2^c,...,s_k^c \rangle \leftarrow \langle \hat{s}_1^c, \hat{s}_2^c,...,\hat{s}_k^c \rangle$ and return Step 2
\STATE {\bf return} $\hat{\mathcal{Z}}$, $\langle \hat{s}_1^c, \hat{s}_2^c,...,\hat{s}_k^c \rangle$
\end{algorithmic}
\end{algorithm}

\section{Experiments}
\subsection{Benchmarks and Implementation Details}
We utilize GPT-3.5 (\texttt{gpt-3.5-turbo}) and GPT-4 (\texttt{gpt-4}) as our LLM agents, and the evaluations are carried out using the Azure OpenAI API. We set the temperature to 0.0 to generate responses via the greedy algorithm. The specific prompts used for each LLM agent are detailed in the Appendix (from Table \ref{tab:semantic-dce-prompt} to Table \ref{tab:paragraph-improver-prompt}). All experiments are conducted on our local machine (Macbook-Pro with M1 chip) without the need for GPU resources. In our experimental setup, we set both $\alpha$ and $\beta$ in Eq.~\ref{eq:amc-final} to 0. We employ four datasets to evaluate \methodName where QQP and PAWS are binary datasets, as well as SummEval and QAGS have numeric scores representing human judgments. 
\vspace{-0.2cm}
\begin{itemize}[leftmargin=10pt]
    \item {\bf QQP and PAWS}: Quora Question Pair corpus \citep{iyer2017qqp} and the Paraphrase Adversaries from Word Scrambling dataset \citep{zhang2019paws} contain pairs of sentences labeled to indicate whether they are paraphrases or not, while PAWS specifically focuses on the adversarial paraphrases. Following the guidance of BERTScore \citep{zhang2019bertscore}, we are using the PAWS development set and the first 5000 from the training set of QQP.
    \item {\bf SummEval} \citep{Fabbri2021summeval} is a standard dataset that assesses various summarization evaluation techniques. It gathers human ratings in various aspects and is built on the CNN/DailyMail dataset \citep{hermann2015teaching}. In this study, we mainly focus on the consistency evaluation.
    \item {\bf QAGS} \citep{wang2020asking} serves as a benchmark for assessing hallucinations in summarization tasks. Its objective is to evaluate the consistency aspect of summaries across two distinct summarization datasets: QGS-CNN and QAGA-XSUM. 
\end{itemize}

{\color{black}Here we provide a detailed explanation of the “reference” used in our experiments. For Paraphrase detection tasks, such as the QQP dataset, each question pair is annotated with a binary value indicating whether the two questions are paraphrases of each other. We consider “question1” as the “reference” and “question2” as the “candidate”, and our task is to evaluate if the candidate is consistent with the reference in semantic meaning. For Summarization tasks, SummEval datasets include original source articles, machine summaries, and human summaries. Our “reference” in this task is the original source article, and our “candidate” is the machine summaries. Our task is to check the factual consistency between them without relying on any additional golden reference or ground truth.}

\subsection{Baseline Methods}

We evaluate \methodName against a variety of evaluation metrics and LLM-based evaluators that have achieved state-of-the-art performance. 

\begin{itemize}[leftmargin=10pt,nosep]
\item {\bf ROUGE} \citep{lin2004rouge} is widely used evaluation metric with three different variants ROUGE-1, ROUGE-2, and ROUGE-L. We are using ROUGE-2 and ROUGE-L as comparisons in our study. 
\item {\bf BERTScore} \citep{zhang2019bertscore} calculates the similarities between two pieces of text using the contextualized embedding derived from the BERT model\citep{devlin2019bert}. It operates as a similarity-based assessment tool, which has been widely used for various applications. 
\item {\bf MoverScore} \citep{zhao2019mover} enhances BERTScore by incorporating soft alignments and introducing new aggregation techniques to provide a more robust similarity assessment.
\item {\bf BARTScore} \citep{yuan2021bartscore} is a comprehensive evaluator that uses the average likelihood of the model's output as its measurement criteria. 
\item {\bf UniEval} \citep{zhong2022towards} is a consolidated evaluator capable of assessing various elements of text generation as QA tasks. It manages diverse evaluation tasks by modifying the question format.
\item {\bf GPTScore} \citep{fu2023gptscore} is an LLM-based evaluator that assesses texts using pre-training models, e.g., GPT-3, and is designed to provide a higher likelihood to high-quality generated text.
\item {\bf G-Eval} \citep{liu2023gpteval} is another LLM evaluator that utilizes LLMs with a chain-of-thoughts (CoT) approach with a form-filling paradigm to evaluate the quality of NLG outputs.
\end{itemize}

\subsection{Main results on Consistency Evaluation (\divideconquer-\metric)}

\paragraph{Semantic Consistency Evaluation.}
Table \ref{tab:qqp} shows the Area Under the ROC curve (AUROC) for automatic baseline metrics and our method,  following the practice of BERTScore \citep{zhang2019bertscore}. We note that while most metrics from BERTScore perform acceptably on QQP, they exhibit a significant performance drop on PAWS. This suggests that these baseline metrics struggle to detect the challenging adversarial examples from a semantic consistency perspective. In contrast, our method, whether implemented with GPT-3.5 or GPT-4, outperforms all the baseline metrics on both QQP and PAWS, without a significant drop. Notably, \divideconquer-\metric-4 demonstrates superior robustness in adversarial paraphrase classification (semantic consistency) achieving a relatively large improvement (+4.6\% in QQP and +9.9\% in PAWS) compared to BERTScore. 

\begin{wraptable}{r}{7cm}
\vspace{-0.4cm}
\caption{AUROC results on QQP and PAWS}
\vspace{-0.3cm}
\label{tab:qqp}
\centering
\small
\begin{tabular}{@{}l|cc@{}}
\toprule
Metrics        & QQP & PAWS \\ \midrule
BLEU \citep{papineni2002bleu}         & 0.707   & 0.527        \\
METEOR \citep{banerjee2005meteor}         & 0.755   & 0.532        \\
ROUGE-L \citep{lin2004rouge}        & 0.740   & 0.536        \\
CHRF++ \citep{popovic2015chrf}        & 0.577   & 0.608        \\
BEER \citep{stanojevic2014beer}           & 0.741   & 0.564        \\
EED \citep{stanchev2019eed}           & 0.743   & 0.611        \\
CharacTER \citep{wang2016character}      & 0.698   & 0.650        \\
BERTScore \citep{zhang2019bertscore}      & 0.777   & 0.693        \\\midrule
\divideconquer-\metric-3.5 (our method)        & 0.788   & 0.770        \\
\divideconquer-\metric-4 (our method)         & {\bf 0.823}   & {\bf 0.792}        \\
\bottomrule
\end{tabular}
\end{wraptable}

\paragraph{Factual Consistency Evaluation.}
While advanced NLG models are capable of generating high-quality responses, LLMs are known to occasionally produce non-factual statements or hallucinate facts, which can undermine trust in their output. Recent work \citep{manakul2023selfcheckgpt} has been conducted to identify such inconsistencies in terms of factuality. To verify the effectiveness of our method in evaluating hallucination, we test it on the QAGS benchmark, which includes two summarization datasets: QAGS-CNN and QAGS-XSUM. Table \ref{tab:qags} provides a comprehensive comparison of various metrics based on Pearson, Spearman, and Kendall-Tau correlations. We observe that BARTScore performs competitively on the extractive subset (QAGS-CNN) but fails to demonstrate a high correlation on the abstractive subset (QAGS-XSUM). UniEval exhibits a better correlation than G-Eval-3.5 but is comparable to G-Eval-4. Our proposed \divideconquer-\metric-4 outperforms all the baseline methods on both subsets, particularly by a significant margin on QAGS-XSUM. Unlike the G-Eval method, which shows a larger gap between GPT-3.5 and GPT-4, our \divideconquer-\metric~method remains relatively stable when switching between LLMs.

It's crucial to note that QAGS-XSUM is an abstractive dataset, and its summaries are typically one sentence long. This contrasts with the extractive database of QAGS-CNN, where summaries are composed of multiple sentences. Consequently, our method operates at a sentence level for QAGS-XSUM, and our final score is always either 0 or 1. Furthermore, the binary label in QAGS-XSUM implies that we achieve the same correlation score using different correlation methods.

\begin{wraptable}{r}{7cm}
\vspace{-0.45cm}
\caption{Correlation ($\rho$ and $\tau$) results of different metrics on SummEval benchmark.}
\vspace{-0.3cm}
\label{tab:summeval}
\centering
\small
\begin{tabular}{@{}c|cc@{}}
\toprule
\multirow{2}{*}{Metrics} & \multicolumn{2}{c}{SummEval-Consistency}                                       \\ \cmidrule(l){2-3} 
                         & Spearman ($\rho$) & Kendall-Tau ($\tau$)\\ \cmidrule(r){1-1}
ROUGE-2                  & 0.187                              & 0.155                                 \\
ROUGE-L                  & 0.115                              & 0.092                                 \\\midrule
BARTScore                & 0.382                              & 0.315                                 \\
BERTScore                & 0.110                              & 0.090                                 \\
MoverScore               & 0.152                              & 0.127                                 \\
UniEval                  & 0.446                              & 0.371                                 \\\midrule
GPT-Score                & 0.449                              & -                                 \\
G-Eval-3.5               & 0.386                              & 0.318                                 \\
G-Eval-4                 & 0.507                              & 0.425                                 \\\midrule
\divideconquer-\metric-3.5                  & 0.592                              & 0.563                                 \\
\divideconquer-\metric-4                    & {\bf 0.700} (+19.3\%)                              & {\bf 0.668} (+24.3\%)                                 \\
\bottomrule
\end{tabular}
\end{wraptable}

\paragraph{Summarization Consistency Evaluation.}
We follow the setting of previous work \citep{zhong2022towards} to evaluate different summarization consistency using summary-level Spearman ($\rho$) and Kendall-Tau ($\tau$) correlation. As shown in Table \ref{tab:summeval}, baseline metrics using semantic similarity, such as ROUGE and BERTScore, perform poorly on consistency evaluations. While LLM-based evaluators like GPT-Score and G-Eval exhibit higher correlations, they still underperformed compared to our proposed method. \divideconquer-\metric-4 achieves much higher human correspondence compared to \divideconquer-\metric-3.5 on both Spearman and Kendall-Tau correlation, which indicates that the larger size of GPT-4 model is beneficial for sentence-level consistency checking in summarization tasks. \divideconquer-\metric-4 with stronger correlations of $\rho=0.7$ and $\tau=0.668$, substantially improves upon the G-Eval-4 baseline by a large margin (+19.3\% and +24.3\% respectively).

\begin{table}[h!]
\caption{Pearson ($r$), Spearman ($\rho$), and Kendall-Tau ($\tau$) correlations of different baseline metrics on QAGS-CNN and QAGS-XSUM benchmark.}
\vspace{-0.3cm}
\label{tab:qags}
\resizebox{\linewidth}{!}{
\centering
\small
\begin{tabular}{@{}c|ccc|ccc@{}}
\toprule
\multirow{2}{*}{Metrics} & \multicolumn{3}{c}{QAGS-CNN}                                                     & \multicolumn{3}{c}{QAGS-XSUM}                                                    \\ \cmidrule(l){2-7} 
                         & Pearson ($r$) & Spearman ($\rho$) & Kendall-Tau ($\tau$) & Pearson ($r$) & Spearman ($\rho$) & Kendall-Tau ($\tau$)  \\ \cmidrule(r){1-1}
ROUGE-2                     & 0.459           & 0.418                              & 0.333                                 & 0.097           & 0.083                              & 0.068                                 \\
ROUGE-L                     & 0.357           & 0.324                              & 0.254                                & 0.024           & -0.011                              & -0.009                                 \\ \midrule

BARTScore                & 0.735           & 0.680                              & 0.557                                 & 0.184           & 0.159                              & 0.130                                 \\
BERTScore                & 0.576           & 0.505                              & 0.399                                 & 0.024           & 0.008                              & 0.006                                 \\

MoverScore               & 0.414           & 0.347                              & 0.271                                 & 0.054           & 0.044                              & 0.036                                 \\

UniEval                  & 0.682           & 0.662                              & 0.532                                 & 0.461           & 0.488                              & 0.399                                 \\ \midrule
G-Eval-3.5               & 0.477           & 0.516                             & 0.410                                & 0.211           & 0.406                             & 0.343                                 \\
G-Eval-4               & 0.631           & 0.685                              & 0.591                                 & 0.558           & 0.537                 & 0.472 
\\\midrule
\divideconquer-\metric-3.5                  & 0.699           & 0.648                              & 0.596                                 & 0.573           & 0.573                              & 0.573                                 \\
\divideconquer-\metric-4                    & {\bf 0.782}           & {\bf 0.760}                              & {\bf 0.706}                                 & {\bf 0.602}           & {\bf 0.602}                             & {\bf 0.602}                                 \\
\bottomrule
\end{tabular}
}
\end{table}

\subsection{Results for Consistency Improvement (\reasoning)}
After implementing \divideconquer~and \metric, we can quantitatively determine whether each candidate is consistent (score = 1) to the reference or not (score <1). Table \ref{tab:consistency} offers a statistical analysis of the number of inconsistent data after evaluations (\divideconquer-\metric), revealing 286, 111, and 86 inconsistent candidates for the SummEval, QAGS-CNN, and QAGS-XSUM benchmarks respectively. Identifying these inconsistent candidates is valuable but the more critical objective is how to improve these responses to align with the references. To achieve this goal, we re-generate a new response by implementing \reasoning~based on the reasons provided by \divideconquer, and then use \divideconquer~to re-evaluate these improved responses. We observe a significant improvement, with most inconsistencies corrected, specifically 84 out of 86 examples on the QAGS-XSUM benchmark. The rate of consistency improvement is $86.71\%$, $88.29\%$, and $97.67\%$ on SummEval, QAGS-CNN, and QAGS-XSUM respectively. These impressive results demonstrate that our reasoning approach \reasoning~not only provides better consistency evaluation metrics that align more closely with human judgments, but also sheds light on improving consistency beyond evaluation. This finding is particularly crucial for mitigating hallucination once we detect non-factual statements via consistency checks. It's worth noting that our reasoning method \reasoning~is a generic component that can also be applied directly at the paragraph level, and the improvement in this context is significant as well, as illustrated in Table \ref{tab:consistency}.

\begin{table}[h!]
\caption{Performance of consistency improvement with \reasoning~on three benchmark datasets. }
\vspace{-0.3cm}
\label{tab:consistency}
\centering
\small
\begin{tabular}{@{}c|cc|cc|cc@{}}
\toprule
Dataset   (size)             & \multicolumn{2}{c}{SummEval (1600)} & \multicolumn{2}{c}{QAGS-CNN (236)} & \multicolumn{2}{c}{QAGS-XSUM (239)} \\ \midrule
                             & Sentence         & Paragraph        & Sentence        & Paragraph        & Sentence         & Paragraph        \\
Inconsistent data            & 286              & 209              & 111             & 68               & 86               & 90               \\
Corrected data with \reasoning  & 248               & 198                & 89              & 64                & 84                & 82                \\ \midrule
Consistency improvement rate & 86.71\%          & 94.73\%          & 88.29\%         & 94.11\%          & 97.67\%          & 91.11\%          \\ \bottomrule
\end{tabular}
\end{table}

\subsection{Analysis}
\begin{figure}[h!]
\centering
    \includegraphics[width=0.32\textwidth]{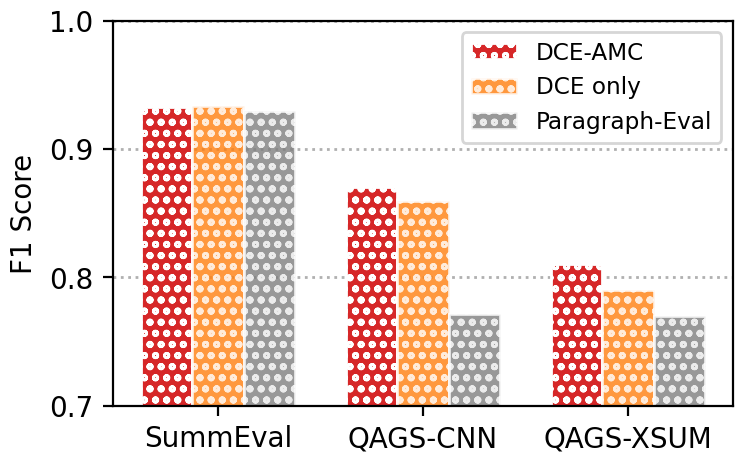}
    \includegraphics[width=0.32\textwidth]{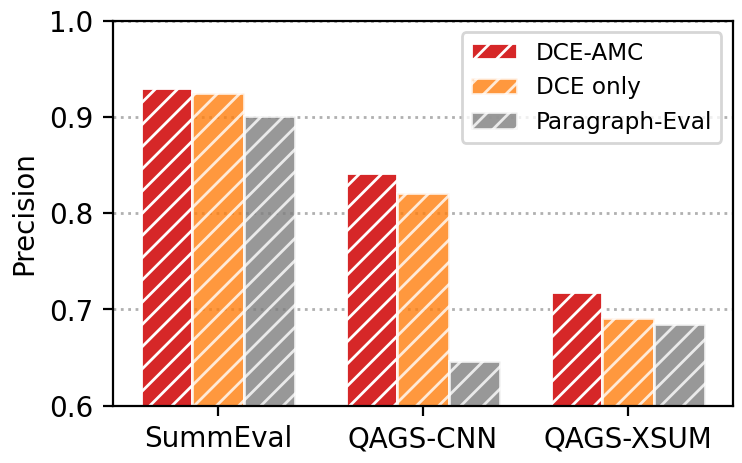}
    \includegraphics[width=0.33\textwidth]{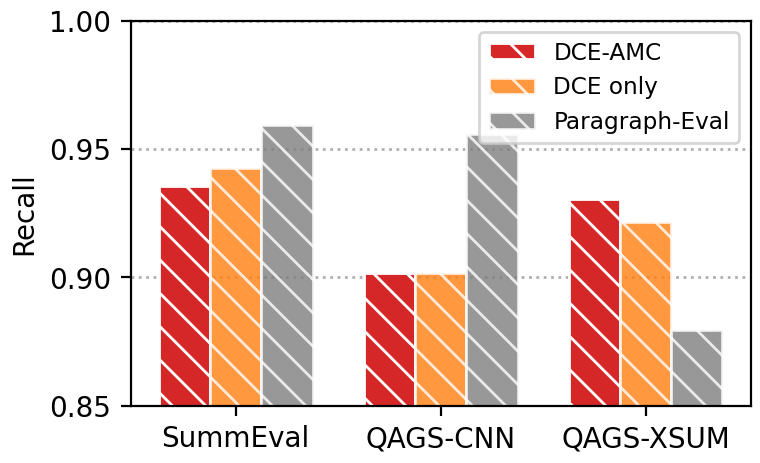}
    \vspace{-0.2cm}
    \caption{F1 score, precision, and recall performance of our method on sentence-level and paragraph-level evaluations. We also verify the effectiveness of the auto-metric converter. }
    \label{fig:sentence-paragraph}
\end{figure}

\paragraph{Why \methodName Prefers Sentence-level Evaluation?} 
To further assess the potential advantage of the sentence-level approach in consistency checking, we employed the same logic of outputting decisions and reasons as used in \divideconquer~ and developed an evaluator at the paragraph level, with prompts provided in Appendix (Table \ref{tab:paragraph_dce_prompt}). The comparative results between paragraph level and sentence level can be viewed in Fig. \ref{fig:sentence-paragraph}. While the recall of paragraph-level evaluation is higher on SummEval and QAGS-CNN benchmarks, its overall performance in terms of the F1 score and precision is lower than that of sentence-level evaluations, particularly on the QAGS benchmark. This combination of higher recall and lower precision implies that more candidates are incorrectly marked as consistent. In the task of consistency checking, a metric with low recall and high precision (sentence level) is preferable as it contributes to higher safety compared to a metric with high recall and low precision (paragraph level), erring on the side of caution. 

In addition to superior accuracy, sentence-level evaluations can facilitate more thorough inconsistency remediation when integrating with \reasoning.  We compared the performance improvement between our sentence level \divideconquer~and paragraph level, as indicated in Table \ref{tab:consistency}. Despite the higher recall of the paragraph-level approach, fewer items are flagged as inconsistent, resulting in fewer candidates being corrected, even though the improvement rate is higher. In fact sentence level \divideconquer~leads to 25.25\% and 39.05\% more corrections compared to the paragraph-level approach in SummEval and QAGS-CNN respectively. Therefore, our sentence-level approach not only outperforms in terms of F1 score and precision during consistency checks, but also facilitates comprehensive improvements through \reasoning.

\paragraph{Is Auto-metric Converter Necessary?} We present a comparison of our method, both with and without \metric, as shown in Fig. \ref{fig:sentence-paragraph}. We observe that our method with only the \divideconquer~\textit{(red bar)} performs marginally better on the SummEval dataset but underperforms \divideconquer-\metric~\textit{(orange bar)} on all other benchmarks. Although \divideconquer~plays a key role in our method, the \metric~component is still desirable and highly necessary not only because it shows better performance, but also because it facilitates the conversion of {\em reasons} outputted by \divideconquer~to a numeric system. This conversion is both user-friendly and practical, making it easy for humans to understand and apply. Furthermore, it provides a straightforward means of evaluating the effectiveness of the \divideconquer~component. 

\vspace{-0.2cm}
\begin{figure}[h!]
\begin{minipage}[t]{0.66\textwidth}
\centering
    \includegraphics[width=4.5cm]{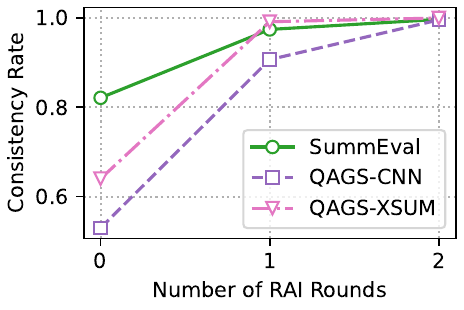}
    \includegraphics[width=4.5cm]{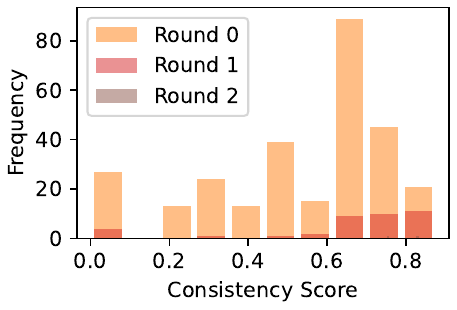}
    \vspace{-0.3cm}
\caption{Multi-round consistency improvement}
\label{fig:multi-round}
\end{minipage}
\begin{minipage}[t]{0.3\textwidth}
\centering
    \includegraphics[width=4.6cm]{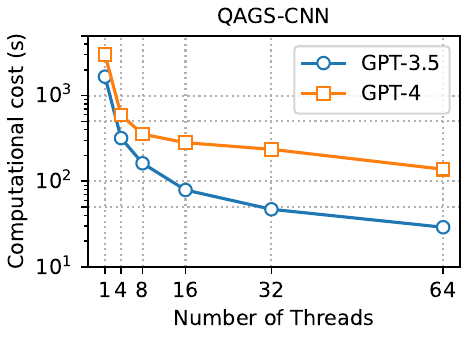}
        \vspace{-0.7cm}
\caption{Computational cost.} 
\label{fig:time}
\end{minipage}
\end{figure}

\paragraph{Multi-round Consistency Improvement.}  
Table \ref{tab:consistency} showcases encouraging results on consistency improvement via \reasoning. This naturally leads to the question: can we further enhance the consistency through multiple rounds of \reasoning?  Fig. \ref{fig:multi-round} shows our investigation on multi-round consistency improvement by iteratively applying \reasoning. It is noteworthy that the convergence of consistency improvement is remarkably swift, achieving nearly 100\% in just two rounds. The convergence rate on the QAGS datasets is highly consistent across both subsets, slightly surpassing SummEval due to its high initial rate after the first round of \reasoning. This is also corroborated by the frequency distribution of the consistency score (Fig. \ref{fig:multi-round} (right)). As the number of rounds increases, the lower consistency scores (<1) gradually decrease, and more inconsistent candidates tend to be consistent, where the score is 1.  

\paragraph{The Effect of LLM models.}
We evaluated the performance of our method using different LLMs across all three benchmarks. It is noteworthy that \divideconquer-\metric-4 generally outperforms \divideconquer-\metric-3.5 across most datasets. The performance gap between the two LLM models is relatively minor in terms of semantic consistency (QQP and PAWS in Table \ref{tab:qqp}), and the abstractive subset (QAGS-XSUM in Table \ref{tab:qags}) in factual consistency, but a significant difference is observed in summarization consistency in Table \ref{tab:summeval}. This suggests that GPT-4 can further enhance performance, especially for more complex evaluation tasks. As such, we applied \reasoning~with GPT-4 directly to verify its superior capability in consistency improvement. Nonetheless, the benefits of GPT-3.5, such as higher computational efficiency and lower API costs, should not be overlooked.

\paragraph{Computational Cost.} We assessed the computational cost of our method based on wall-clock time, which is primarily consumed by LLMs inference. However, the divide-conquer strategy we employed is scalable and easily implemented in parallel. Fig.~\ref{fig:time} illustrates the computational cost of GPT-3.5 and GPT-4 with varying numbers of threads on the QAGS-CNN benchmark. A clear reduction in computational cost is observed as the number of threads increases. It's important to note that the decrease in time is more significant when transitioning from a single thread to four threads, but tends to plateau as more threads are utilized. While GPT-3.5, being the smaller LLM, is a more efficient option, GPT-4 often delivers better performance.

\section{Related Work}
\paragraph{LLM-based Evaluations.} Unlike conventional evaluating metrics leveraging token-level or similarity embeddings, such as BLEU \citep{papinenibleu2002}, ROUGE \citep{lin2004rouge}, or BERTScore \citep{zhang2019bertscore}, recent proposed LLM-based evaluators \citep{wang2023chatgpt}, such as GPTScore \citep{fu2023gptscore} and G-Eval \citep{liu2023gpteval}, have demonstrated competitive performance on multiple NLG tasks. Their idea is to utilize the LLMs to score the candidate output under the assumption that the LLMs have learned to assign higher probabilities to fluent and high-quality contexts. However, these LLM evaluators often exhibit lower correlations with human judgments, and their reliability, robustness, and validity remain under-explored \citep{liu2023gpteval}. Specifically, LLM evaluators may pose potential risks of producing hallucinated or overconfidence scores if the LLM model is not well calibrated for complex tasks \citep{kadavath2022language, zhou2023navigating}. This results in limited confidence in using LLM evaluators to directly evaluate paragraph-level responses. Our proposed \methodName framework addresses these challenges through a divide-conquer strategy (\divideconquer) coupled with a numeric score system (\metric). Our method quantitatively evaluates paragraphs sentence-by-sentence and does not rely on LLMs to directly output numeric scores, thus providing a more accurate and comprehensive score that better aligns with human feedback. 

\paragraph{Consistency Evaluations.} 
Consistency checking plays an essential role in a wide range of NLG tasks, including question-answering \citep{durmus2020, wang2020asking}, factual knowledge extraction \citep{elazar2021measuring}, summarization \citep{durmus2020} and hallucination detection \citep{manakul2023selfcheckgpt}. However, due to various limitations of existing methods, such as reliance on additional pre-trained models or question sets \citep{durmus2020}, it is highly desirable to develop a unified and automatic consistency metric \citep{wang2022self}. Our proposed framework successfully fills this gap and demonstrates superior performance compared to state-of-the-art baselines \citep{fu2023gptscore,liu2023gpteval,wang2023chatgpt}. More importantly, our proposed \reasoning~enables consistency improvement where the re-generated candidate response significantly helps mitigate LLM hallucinations \citep{dhuliawala2023chain, mundler2023self, zhang2023language} in summarization, and open-book QA tasks \citep{li2023halueval}.   

\section{Conclusion and Discussion}
We proposed a general evaluation framework based on a divide-and-conquer strategy for assessing the consistency between the LLM-generated output and the reference texts across various NLG tasks. Moreover, the proposed method can leverage analytical reasoning to generate revised text with improved consistency. Through comprehensive and systematic empirical study across multiple benchmarks in semantic, factual, and summarization consistency tasks, we demonstrated that our approach significantly outperforms existing methods in evaluating and enhancing the consistency of LLM-generated content. Despite these advancements, we acknowledge several potential limitations of our proposed method:

{\bf Not a Silver Bullet.}
While our sentence-level approach (\divideconquer-\metric) excels in evaluating {\em consistency} and {\em detecting hallucination}, it may not be universally effective for all dimensions of text evaluation, even with updated criteria in prompts. For instance, dimensions such as \textit{coherence}, which pertains to the collective quality of all generated sentences, or \textit{relevance}, which involves selecting important information and eliminating redundant content from the reference text, require a holistic focus on the entire candidate. These dimensions may not be ideally suited for our \divideconquer-\metric~approach. However, if a different evaluator that outputs reasons for action is used, our \metric~and \reasoning~could still be employed to quantify and improve performance on such dimensions.

{\bf Garbage in, Garbage Out.} The \methodName framework requires two inputs: a reference paragraph and a candidate paragraph. As we use the reference paragraph as the target for consistency and hallucination checks, any non-factual statements present in the reference paragraph would not be detected by our method. Therefore, for tasks such as retrieval-augmented generation (RAG), the accuracy of our method is inherently limited by the correctness of the input paragraphs.

{\color{black}{\bf Meta-prompting.} Our DCR framework requires hand-craft prompts for specific tasks, and acknowledges that this is a general hurdle shared by all works relying on LLMs, which include G-Eval \citep{liu2023gpteval}, GPTScore \citep{fu2023gptscore}, and Self-refine \citep{madaan2023self}. Specifically, in G-Eval, different prompts will need to be composed for different aspects: consistency, coherence, etc. Self-refine defines multiple customized prompts to perform their INIT - FEEDBACK – REFINE components. Our current solution is to structure our prompts in a modularized manner so task-specific content can be updated easily. However, an automated prompt-tuning procedure is beyond the focus of our study but we leave this for future work.}

\bibliography{iclr2024_conference}

\begin{thebibliography}{46}
\providecommand{\natexlab}[1]{#1}
\providecommand{\url}[1]{\texttt{#1}}
\expandafter\ifx\csname urlstyle\endcsname\relax
  \providecommand{\doi}[1]{doi: #1}\else
  \providecommand{\doi}{doi: \begingroup \urlstyle{rm}\Url}\fi

\bibitem[amd Jason~Baldridge \& He(2019)amd Jason~Baldridge and
  He]{zhang2019paws}
Yuan~Zhang amd Jason~Baldridge and Luheng He.
\newblock Paws: Paraphrase adversaries from word scrambling.
\newblock \emph{arXiv preprint arXiv:1904.01130}, 2019.

\bibitem[Amplayo et~al.(2022)Amplayo, Liu, Zhao, and Narayan]{amplayo2022smart}
Reinald~Kim Amplayo, Peter~J Liu, Yao Zhao, and Shashi Narayan.
\newblock Smart: Sentences as basic units for text evaluation.
\newblock In \emph{The Eleventh International Conference on Learning
  Representations}, 2022.

\bibitem[Banerjee \& Lavie(2005)Banerjee and Lavie]{banerjee2005meteor}
Satanjeev Banerjee and Alon Lavie.
\newblock Meteor: An automatic metric for mt evaluation with improved
  correlation with human judgments.
\newblock In \emph{Proceedings of the acl workshop on intrinsic and extrinsic
  evaluation measures for machine translation and/or summarization}, pp.\
  65--72, 2005.

\bibitem[Bubeck et~al.(2023)Bubeck, Chandrasekaran, Eldan, Gehrke, Horvitz,
  Kamar, Lee, Lee, Li, Lundberg, et~al.]{bubeck2023sparks}
S{\'e}bastien Bubeck, Varun Chandrasekaran, Ronen Eldan, Johannes Gehrke, Eric
  Horvitz, Ece Kamar, Peter Lee, Yin~Tat Lee, Yuanzhi Li, Scott Lundberg,
  et~al.
\newblock Sparks of artificial general intelligence: Early experiments with
  gpt-4.
\newblock \emph{arXiv preprint arXiv:2303.12712}, 2023.

\bibitem[Chen et~al.(2023)Chen, Phang, Parrish, Padmakumar, Zhao, Bowman, and
  Cho]{chen2023two}
Angelica Chen, Jason Phang, Alicia Parrish, Vishakh Padmakumar, Chen Zhao,
  Samuel~R Bowman, and Kyunghyun Cho.
\newblock Two failures of self-consistency in the multi-step reasoning of llms.
\newblock \emph{arXiv preprint arXiv:2305.14279}, 2023.

\bibitem[Devlin et~al.(2019)Devlin, Chang, aLee, and Toutanova]{devlin2019bert}
Jacob Devlin, Ming-Wei Chang, Kenton aLee, and Kristina Toutanova.
\newblock Bert: Pre-training of deep bidirectional transformers for language
  understanding.
\newblock \emph{arXiv preprint arXiv:1810.04805}, 2019.

\bibitem[Dhuliawala et~al.(2023)Dhuliawala, Komeili, Xu, Raileanu, Li,
  Celikyilmaz, and Weston]{dhuliawala2023chain}
Shehzaad Dhuliawala, Mojtaba Komeili, Jing Xu, Roberta Raileanu, Xian Li, Asli
  Celikyilmaz, and Jason Weston.
\newblock Chain-of-verification reduces hallucination in large language models.
\newblock \emph{arXiv preprint arXiv:2309.11495}, 2023.

\bibitem[Durmus et~al.(2020)Durmus, He, and Diab]{durmus2020}
Esin Durmus, He~He, and Mona Diab.
\newblock Feqa: A question answering evaluation framework for faithfulness
  assessment in abstractive summarization.
\newblock \emph{arXiv preprint arXiv:2005.03754}, 2020.

\bibitem[Elazar et~al.(2021)Elazar, Kassner, Ravfogel, Ravichander, Hovy,
  Sch{\"u}tze, and Goldberg]{elazar2021measuring}
Yanai Elazar, Nora Kassner, Shauli Ravfogel, Abhilasha Ravichander, Eduard
  Hovy, Hinrich Sch{\"u}tze, and Yoav Goldberg.
\newblock Measuring and improving consistency in pretrained language models.
\newblock \emph{Transactions of the Association for Computational Linguistics},
  9:\penalty0 1012--1031, 2021.

\bibitem[Fabbri et~al.(2021)Fabbri, Kryscinski, McCann, Xiong, Socher, and
  Radev]{Fabbri2021summeval}
Alexander~R. Fabbri, Wojciech Kryscinski, Bryan McCann, Caiming Xiong, Richard
  Socher, and Dragomir Radev.
\newblock Summeval: Re-evaluating summarization evaluation.
\newblock \emph{arXiv preprint arXiv:2007.12626}, 2021.

\bibitem[Hanna \& Bojar(2021)Hanna and Bojar]{hanna2021fine}
Michael Hanna and Ond{\v{r}}ej Bojar.
\newblock A fine-grained analysis of bertscore.
\newblock In \emph{Proceedings of the Sixth Conference on Machine Translation},
  pp.\  507--517, 2021.

\bibitem[Hermann et~al.(2015)Hermann, Kocisky, Grefenstette, Espeholt, Kay,
  Suleyman, and Blunsom]{hermann2015teaching}
Karl~Moritz Hermann, Tomas Kocisky, Edward Grefenstette, Lasse Espeholt, Will
  Kay, Mustafa Suleyman, and Phil Blunsom.
\newblock Teaching machines to read and comprehend.
\newblock \emph{Advances in neural information processing systems}, 28, 2015.

\bibitem[Iyer et~al.(2017)Iyer, Dandekar, and Csernai]{iyer2017qqp}
Shankar Iyer, Nikhil Dandekar, and Kornel Csernai.
\newblock First quora dataset release: Question pairs.
\newblock 2017.

\bibitem[Ji et~al.(2023)Ji, Lee, Frieske, Yu, Su, Xu, Ishii, Bang, Madotto, and
  Fung]{ji2023survey}
Ziwei Ji, Nayeon Lee, Rita Frieske, Tiezheng Yu, Dan Su, Yan Xu, Etsuko Ishii,
  Ye~Jin Bang, Andrea Madotto, and Pascale Fung.
\newblock Survey of hallucination in natural language generation.
\newblock \emph{ACM Computing Surveys}, 55\penalty0 (12):\penalty0 1--38, 2023.

\bibitem[Jiang et~al.(2023)Jiang, Ren, and Lin]{jiang2023llm}
Dongfu Jiang, Xiang Ren, and Bill~Yuchen Lin.
\newblock Llm-blender: Ensembling large language models with pairwise ranking
  and generative fusion.
\newblock \emph{arXiv preprint arXiv:2306.02561}, 2023.

\bibitem[Jinlan et~al.(2023)Jinlan, See-Kiong, Zhengbao, and
  Pengfei]{fu2023gptscore}
Fu~Jinlan, Ng~See-Kiong, Jiang Zhengbao, and Liu Pengfei.
\newblock Gptscore: Evaluate as you desire.
\newblock \emph{arXiv preprint arXiv:2302.04166}, 2023.

\bibitem[Kadavath et~al.(2022)Kadavath, Conerly, Askell, Henighan, Drain,
  Perez, Schiefer, Hatfield-Dodds, DasSarma, Tran-Johnson,
  et~al.]{kadavath2022language}
Saurav Kadavath, Tom Conerly, Amanda Askell, Tom Henighan, Dawn Drain, Ethan
  Perez, Nicholas Schiefer, Zac Hatfield-Dodds, Nova DasSarma, Eli
  Tran-Johnson, et~al.
\newblock Language models (mostly) know what they know.
\newblock \emph{arXiv preprint arXiv:2207.05221}, 2022.

\bibitem[Kamalloo et~al.(2023)Kamalloo, Dziri, Clarke, and
  Rafiei]{kamalloo2023evaluating}
Ehsan Kamalloo, Nouha Dziri, Charles~LA Clarke, and Davood Rafiei.
\newblock Evaluating open-domain question answering in the era of large
  language models.
\newblock \emph{arXiv preprint arXiv:2305.06984}, 2023.

\bibitem[Kuhn et~al.(2023)Kuhn, Gal, and Farquhar]{kuhn2023semantic}
Lorenz Kuhn, Yarin Gal, and Sebastian Farquhar.
\newblock Semantic uncertainty: Linguistic invariances for uncertainty
  estimation in natural language generation.
\newblock \emph{arXiv preprint arXiv:2302.09664}, 2023.

\bibitem[Lewis et~al.(2020)Lewis, Perez, Piktus, Petroni, Karpukhin, Goyal,
  K{\"u}ttler, Lewis, Yih, Rockt{\"a}schel, et~al.]{lewis2020retrieval}
Patrick Lewis, Ethan Perez, Aleksandra Piktus, Fabio Petroni, Vladimir
  Karpukhin, Naman Goyal, Heinrich K{\"u}ttler, Mike Lewis, Wen-tau Yih, Tim
  Rockt{\"a}schel, et~al.
\newblock Retrieval-augmented generation for knowledge-intensive nlp tasks.
\newblock \emph{Advances in Neural Information Processing Systems},
  33:\penalty0 9459--9474, 2020.

\bibitem[Li et~al.(2023)Li, Cheng, Zhao, Nie, and Wen]{li2023halueval}
Junyi Li, Xiaoxue Cheng, Wayne~Xin Zhao, Jian-Yun Nie, and Ji-Rong Wen.
\newblock Halueval: A large-scale hallucination evaluation benchmark for large
  language models.
\newblock \emph{arXiv e-prints}, pp.\  arXiv--2305, 2023.

\bibitem[Lin(2004)]{lin2004rouge}
Chin-Yew Lin.
\newblock Rouge: A package for automatic evaluation of summaries.
\newblock In \emph{Text summarization branches out}, pp.\  74--81, 2004.

\bibitem[Liu et~al.(2023{\natexlab{a}})Liu, Jin, Wang, Cheng, Dou, and
  Wen]{liu2023reta}
Jiongnan Liu, Jiajie Jin, Zihan Wang, Jiehan Cheng, Zhicheng Dou, and Ji-Rong
  Wen.
\newblock Reta-llm: A retrieval-augmented large language model toolkit.
\newblock \emph{arXiv preprint arXiv:2306.05212}, 2023{\natexlab{a}}.

\bibitem[Liu et~al.(2023{\natexlab{b}})Liu, Iter, Xu, Wang, Xu, and
  Zhu]{liu2023gpteval}
Yang Liu, Dan Iter, Yichong Xu, Shuohang Wang, Ruochen Xu, and Chenguang Zhu.
\newblock Gpteval: Nlg evaluation using gpt-4 with better human alignment.
\newblock \emph{arXiv preprint arXiv:2303.16634}, 2023{\natexlab{b}}.

\bibitem[Madaan et~al.(2023)Madaan, Tandon, Gupta, Hallinan, Gao, Wiegreffe,
  Alon, Dziri, Prabhumoye, Yang, et~al.]{madaan2023self}
Aman Madaan, Niket Tandon, Prakhar Gupta, Skyler Hallinan, Luyu Gao, Sarah
  Wiegreffe, Uri Alon, Nouha Dziri, Shrimai Prabhumoye, Yiming Yang, et~al.
\newblock Self-refine: Iterative refinement with self-feedback.
\newblock \emph{arXiv preprint arXiv:2303.17651}, 2023.

\bibitem[Manakul et~al.(2023)Manakul, Liusie, and
  Gales]{manakul2023selfcheckgpt}
Potsawee Manakul, Adian Liusie, and Mark~JF Gales.
\newblock Selfcheckgpt: Zero-resource black-box hallucination detection for
  generative large language models.
\newblock \emph{arXiv preprint arXiv:2303.08896}, 2023.

\bibitem[M{\"u}ndler et~al.(2023)M{\"u}ndler, He, Jenko, and
  Vechev]{mundler2023self}
Niels M{\"u}ndler, Jingxuan He, Slobodan Jenko, and Martin Vechev.
\newblock Self-contradictory hallucinations of large language models:
  Evaluation, detection and mitigation.
\newblock \emph{arXiv preprint arXiv:2305.15852}, 2023.

\bibitem[Papineni et~al.(2002{\natexlab{a}})Papineni, Roukos, Ward, and
  Zhu]{papineni2002bleu}
Kishore Papineni, Salim Roukos, Todd Ward, and Wei-Jing Zhu.
\newblock Bleu: a method for automatic evaluation of machine translation.
\newblock In \emph{Proceedings of the 40th annual meeting of the Association
  for Computational Linguistics}, pp.\  311--318, 2002{\natexlab{a}}.

\bibitem[Papineni et~al.(2002{\natexlab{b}})Papineni, Roukos, Ward, and
  Zhu]{papinenibleu2002}
Kishore Papineni, Salim Roukos, Todd Ward, and WeiJing Zhu.
\newblock Bleu: a method for automatic evaluation of machine translation.
\newblock \emph{Proceedings of the 40th annual meeting of the Association for
  Computational Linguistics, pages 311–318.}, 2002{\natexlab{b}}.

\bibitem[Popovi{\'c}(2015)]{popovic2015chrf}
Maja Popovi{\'c}.
\newblock chrf: character n-gram f-score for automatic mt evaluation.
\newblock In \emph{Proceedings of the tenth workshop on statistical machine
  translation}, pp.\  392--395, 2015.

\bibitem[Rawte et~al.(2023)Rawte, Sheth, and Das]{rawte2023survey}
Vipula Rawte, Amit Sheth, and Amitava Das.
\newblock A survey of hallucination in large foundation models.
\newblock \emph{arXiv preprint arXiv:2309.05922}, 2023.

\bibitem[Stanchev et~al.(2019)Stanchev, Wang, and Ney]{stanchev2019eed}
Peter Stanchev, Weiyue Wang, and Hermann Ney.
\newblock Eed: Extended edit distance measure for machine translation.
\newblock In \emph{Proceedings of the Fourth Conference on Machine Translation
  (Volume 2: Shared Task Papers, Day 1)}, pp.\  514--520, 2019.

\bibitem[Stanojevi{\'c} \& Sima’an(2014)Stanojevi{\'c} and
  Sima’an]{stanojevic2014beer}
Milo{\v{s}} Stanojevi{\'c} and Khalil Sima’an.
\newblock Beer: Better evaluation as ranking.
\newblock In \emph{Proceedings of the Ninth Workshop on Statistical Machine
  Translation}, pp.\  414--419, 2014.

\bibitem[Tam et~al.(2022)Tam, Mascarenhas, Zhang, Kwan, Bansal, and
  Raffel]{tam2022evaluating}
Derek Tam, Anisha Mascarenhas, Shiyue Zhang, Sarah Kwan, Mohit Bansal, and
  Colin Raffel.
\newblock Evaluating the factual consistency of large language models through
  summarization.
\newblock \emph{arXiv preprint arXiv:2211.08412}, 2022.

\bibitem[Wang et~al.(2020)Wang, Cho, and Lewis]{wang2020asking}
Alex Wang, Kyunghyun Cho, and Mike Lewis.
\newblock Asking and answering questions to evaluate the factual consistency of
  summaries.
\newblock \emph{arXiv preprint arXiv:2004.04228}, 2020.

\bibitem[Wang et~al.(2023)Wang, Liang, Meng, Shi, Li, Xu, Qu, and
  Zhou]{wang2023chatgpt}
Jiaan Wang, Yunlong Liang, Fandong Meng, Haoxiang Shi, Zhixu Li, Jinan Xu,
  Jianfeng Qu, and Jie Zhou.
\newblock Is chatgpt a good nlg evaluator? a preliminary study.
\newblock \emph{arXiv preprint arXiv:2303.04048}, 2023.

\bibitem[Wang et~al.(2016)Wang, Peter, Rosendahl, and Ney]{wang2016character}
Weiyue Wang, Jan-Thorsten Peter, Hendrik Rosendahl, and Hermann Ney.
\newblock Character: Translation edit rate on character level.
\newblock In \emph{Proceedings of the First Conference on Machine Translation:
  Volume 2, Shared Task Papers}, pp.\  505--510, 2016.

\bibitem[Wang et~al.(2022)Wang, Wei, Schuurmans, Le, Chi, Narang, Chowdhery,
  and Zhou]{wang2022self}
Xuezhi Wang, Jason Wei, Dale Schuurmans, Quoc Le, Ed~Chi, Sharan Narang,
  Aakanksha Chowdhery, and Denny Zhou.
\newblock Self-consistency improves chain of thought reasoning in language
  models.
\newblock \emph{arXiv preprint arXiv:2203.11171}, 2022.

\bibitem[Yang et~al.(2023)Yang, Jin, Tang, Han, Feng, Jiang, Yin, and
  Hu]{yang2023harnessing}
Jingfeng Yang, Hongye Jin, Ruixiang Tang, Xiaotian Han, Qizhang Feng, Haoming
  Jiang, Bing Yin, and Xia Hu.
\newblock Harnessing the power of llms in practice: A survey on chatgpt and
  beyond.
\newblock \emph{arXiv preprint arXiv:2304.13712}, 2023.

\bibitem[Yuan et~al.(2021)Yuan, Neubig, and Liu]{yuan2021bartscore}
Weizhe Yuan, Graham Neubig, and Pengfei Liu.
\newblock Bartscore: Evaluating generated text as text generation.
\newblock \emph{Advances in Neural Information Processing Systems},
  34:\penalty0 27263--27277, 2021.

\bibitem[Zhang et~al.(2023{\natexlab{a}})Zhang, Li, Das, Malin, and
  Kumar]{zhang2023sac}
Jiaxin Zhang, Zhuohang Li, Kamalika Das, Bradley~A Malin, and Sricharan Kumar.
\newblock Sac $^3$: Reliable hallucination detection in black-box language
  models via semantic-aware cross-check consistency.
\newblock In \emph{The 2023 Conference on Empirical Methods in Natural Language
  Processing}, 2023{\natexlab{a}}.

\bibitem[Zhang et~al.(2023{\natexlab{b}})Zhang, Press, Merrill, Liu, and
  Smith]{zhang2023language}
Muru Zhang, Ofir Press, William Merrill, Alisa Liu, and Noah~A Smith.
\newblock How language model hallucinations can snowball.
\newblock \emph{arXiv preprint arXiv:2305.13534}, 2023{\natexlab{b}}.

\bibitem[Zhang et~al.(2020)Zhang, Kishore, Wu, Weinberger, and
  Artzi]{zhang2019bertscore}
Tianyi Zhang, Varsha Kishore, Felix Wu, Kilian~Q Weinberger, and Yoav Artzi.
\newblock Bertscore: Evaluating text generation with bert.
\newblock \emph{arXiv preprint arXiv:1904.09675}, 2020.

\bibitem[Zhao et~al.(2019)Zhao, Peyrard, Liu, Gao, Meyer, and
  Eger]{zhao2019mover}
Wei Zhao, Maxime Peyrard, Fei Liu, Yang Gao, Christian~M. Meyer, and Steffen
  Eger.
\newblock Moverscore: Text generation evaluating with contextualized embeddings
  and earth mover distance.
\newblock \emph{arXiv preprint arXiv:1909.02622}, 2019.

\bibitem[Zhong et~al.(2022)Zhong, Liu, Yin, Mao, Jiao, Liu, Zhu, Ji, and
  Han]{zhong2022towards}
Ming Zhong, Yang Liu, Da~Yin, Yuning Mao, Yizhu Jiao, Pengfei Liu, Chenguang
  Zhu, Heng Ji, and Jiawei Han.
\newblock Towards a unified multi-dimensional evaluator for text generation.
\newblock \emph{arXiv preprint arXiv:2210.07197}, 2022.

\bibitem[Zhou et~al.(2023)Zhou, Jurafsky, and Hashimoto]{zhou2023navigating}
Kaitlyn Zhou, Dan Jurafsky, and Tatsunori Hashimoto.
\newblock Navigating the grey area: Expressions of overconfidence and
  uncertainty in language models.
\newblock \emph{arXiv preprint arXiv:2302.13439}, 2023.

\end{thebibliography}
\bibliographystyle{iclr2024_conference}

\clearpage

\appendix
\section{Appendix}

\subsection{Improvement Iterations Example}

\begin{table}[h]
\centering
\begin{center}
\caption{SummEval Improvement Iteration Example}
\label{tab:improvement-example}
\begin{tabular}{ m{1.5cm} | m{12.5cm}}
\toprule
 \textbf{Original Article} & Usain Bolt will compete at the IAAF/BTC World Relays in the Bahamas next month , the Jamaica Athletics Administrative Association has announced . The six-time Olympic gold medallist will compete at the relay championship on May 2 and 3 as part of the Jamaican team . ' I 'm happy to be part of the Jamaican team for the IAAF / BTC World Relays in the Bahamas . I am fit , healthy and ready to run , ' said Bolt . \hl{Usain Bolt has confirmed he will be part of Jamaica 's team at the World Relays in the Bahamas Bolt reacts as he wins 4x100m gold at the London Olympic Games in 2012 ' I hear the meet was a lot of fun last year and there was a great atmosphere . Jamaica has a long and successful tradition in relays and when we put on the national colours we always do our best to make the country proud , ' he added . JAAA General Secretary Garth Gayle commented , 'We were extremely pleased that Usain was available for selection and that the world 's fastest man will be running for Jamaica . We can expect some sprint magic on the track in the Bahamas on 2nd and 3rd May . '} ...  \\ 
 \midrule
 \textbf{Original Summary} & Usain bolt will compete at the relay championship on may 2 and 3 as part of the jamaican team . The six-time olympic gold medalist will be part of jamaica 's team at the iaaf/btc world . \hl{Bolt is the iaaf/btc general secretary garth gayle said he always does ` best to make his country proud ' .}  ...  \\ 
 \midrule
 \textbf{Divide-Conquer Evaluator Evaluation Reasons} & \begin{itemize} [leftmargin=3mm]  
  \item This sentence is consistent with the article. The article states that Usain Bolt will compete at the IAAF/BTC World Relays in the Bahamas next month as part of the Jamaican team
  \item This sentence is consistent with the article. The article confirms that Usain Bolt, a six-time Olympic gold medallist, will be part of Jamaica's team at the IAAF/BTC World Relays
   \item \hl{This sentence is not consistent with the article. The article does not state that Bolt is the IAAF/BTC General Secretary. The General Secretary is Garth Gayle. However, the article does mention that Bolt always does his best to make his country proud} 
  \end{itemize} ...\\
   \midrule
 \textbf{First Round Improved Answer} &  Usain bolt will compete at the relay championship on may 2 and 3 as part of the jamaican team. The six-time olympic gold medalist will be part of jamaica 's team at the iaaf/btc world. \hl{Bolt, as stated by the IAAF/BTC General Secretary Garth Gayle, always does his best to make his country proud.} ... \\
    \midrule
  \textbf{Divide-Conquer Evaluator Re-Evaluation Reasons} & \begin{itemize} [leftmargin=3mm]  
  \item This sentence is consistent with the article. The article states that Usain Bolt will compete at the IAAF/BTC World Relays in the Bahamas next month, which is the relay championship mentioned in the summary. The dates and his participation as part of the Jamaican team are also mentioned in the article
  \item This sentence is consistent with the article. The article mentions that Usain Bolt, a six-time Olympic gold medallist, will be part of the Jamaican team at the IAAF/BTC World Relays.
   \item \hl{This sentence is not consistent with the article. The article does mention that Bolt always does his best to make his country proud, but it is Bolt himself who says this, not the IAAF/BTC General Secretary Garth Gayle.'}
  \end{itemize}  ...\\
  \midrule
 \textbf{Second Round Improved Answer} &  Usain bolt will compete at the relay championship on may 2 and 3 as part of the jamaican team. The six-time olympic gold medalist will be part of jamaica 's team at the iaaf/btc world. \hl{Bolt, as stated by himself, always does his best to make his country proud.} ... \\
\bottomrule
\end{tabular}
\end{center}
\end{table}
\clearpage

\subsection{Full sets of Prompts}


\begin{table}[h]
\centering
\caption{Semantic Consistency Divide-Conquer Evaluator Prompt}
\label{tab:semantic-dce-prompt}
\begin{tabular}{p{13cm}}
\toprule
You are an evaluator. You will be given a true answer and an attempt answer. \\
The true answer is the ground truth answer. Attempt answer is the answer you want to evaluate. \\
\\

\textbf{Criteria:} \\
Your task is to evaluate whether the attempt answer is consistent with the true answer. You will evaluate it by: \\ \\
* Listing all the aspects in the attempt answer  \\
* Compare if each aspect exists in the true answer \\
* If it does, compare if the information in the attempt answer is consistent with what is in the true answer \\
* It is OK that not all information from the true answer exists in attempt answer \\
 \\
\textbf{Given:}\\
\#\# True Answer \#\# \\
\{\textit{true answer}\} \\
\#\# Attempt Answer \#\# \\
\{\textit{answer to evaluate}\} \\
\\
\textbf{Task}\\
Work in a step-by-step way to make sure we get the right answer. You will format the output in JSON as follows: \\
\{"reason": [\{"sentence": "original sentence", "reason": "why this sentence is or is not consistent with the true answer"\}],  "is\_consistent": true/false\} \\
\\
Here is the evaluation in JSON format:
\\
\bottomrule
\end{tabular}
\end{table}

\begin{table}[h]
\centering
\caption{Summarization Consistency Divide-Conquer Evaluator Prompt}
\label{tab:summarization-dce-prompt}
\begin{tabular}{p{13cm}}
\toprule
You are an evaluator. You will be given an article and a summary.  \\
The summary contains a summarized version of the article. \\
\\
\textbf{Criteria:} \\
Your task is to evaluate whether the summary is consistent with the article. You will evaluate it by going through each sentence of the summary and check against the following procedures:  \\ \\
* Understands all the aspects in the sentence, who is doing what at when and where and what are the impact etc.  \\
* Compare if each aspect exists in the article  \\
* If it does, compare if the information in this sentence is consistent with what is in the article  \\
* Compare if all the information in this sentence can be directly inferred or entailed from what is in the article, including but not limited to who, what, when, where, etc.  \\
* It is OK that not all information from the article exists in this summary \\
 \\
\textbf{Given:}\\
\#\# Article \#\# \\
\{\textit{article}\} \\
\#\# Summary \#\# \\
\{\textit{summary}\} \\
\\
\textbf{Task} \\
Work in a step-by-step way to make sure we get the right answer. You will format the output in JSON as follows: \\
\{"reason": [\{"sentence": "original sentence", "reason": "why this sentence is or is not consistent with the article. You should start with \"this sentence is consistent with the article\" or \"this sentence is not consistent with the article\" "\}],  "is\_consistent": true/false\} \\
\\
Here is the evaluation in JSON format:
\\
\bottomrule
\end{tabular}
\end{table}

\begin{table}[h]
\centering
\caption{Auto-Metric Converter Prompt}
\label{tab:metric-prompt}
\begin{tabular}{p{13cm}}
\toprule
You are an evaluator. You will be given a list of paragraphs about "attempt answer". Your job is to: \\
* Identify whether each paragraph is positive or negative \\
* If the paragraph is positive, mark it as 1, \\
* If the paragraph is negative, mark it as -1. \\
* Output the mark for each paragraph  in a JSON array \\
 \\
\# Example \\
\textbf{Given paragraphs:} \\
*"The attempt answer is incorrect as it states that employees in the US are not eligible to participate in the ESPP, which contradicts the true answer. So it is incorrect",   \\
*"The attempt answer adds a new aspect that is not in the true answer.", \\ 
*"Yet it does list the correct article. And that is helpful." \\
\\
\textbf{Thought:}  \\
The first paragraph is negative as it mentions the attempt answer is wrong. Thus mark -1 \\
The second paragraph is negative as it adds something that is not in true answer. Thus mark -1 \\
The third paragraph is positive. Thus mark 1 \\
\\
\textbf{Answer:} \\
\{"reason": ["The first paragraph is negative as it mentions the attempt answer is wrong. Thus mark -1", "The second paragraph is negative as it adds something that is not in the true answer. Thus mark -1", "The third paragraph is positive. Thus mark +1"], "answer": [ -1, -1, 1]\} \\
\\
\textbf{Given:}\\
\#\# Attempt Answer \#\#: \\
\{\textit{attempt answer}\} \\
\\
\textbf{Answer:}
\\
\bottomrule
\end{tabular}
\end{table}

\begin{table}[h]
\centering
\caption{Reason-Assisted Improver Prompt}
\label{tab:improver-prompt}
\begin{tabular}{p{13cm}}
\toprule
You are a good writer. You will be given: \\
* An article \\
* A list of objects, each have two fields: sentence and reason \\
  \quad  ** sentence: These sentences are summaries of the given article.  \\
  \quad  ** reason: These are the reasons why the sentence is consistent with the article or not.\\
    \\
Your job is to rewrite these sentences: \\
* If the sentence is consistent with the article, you can keep it as it is \\
* If the sentence is not consistent with the article, you can re-write it to make it consistent with the article based on the reasons given. \\
\\
\textbf{Article}\\
\{\textit{article}\}\\
\textbf{Sentences} \\
\{\textit{sentences}\} \\
\\
\textbf{Task} \\
Work in a step-by-step way to make sure we get the right answer. You will format the output in JSON as follows: \\
\dummy{[ \{"sentence": "original sentence", "improved\_sentence": "improved sentence", "reason": "if it is improved, how it is improved. if not, say 'ALREADY CONSISTENT'"\}]}

\\
\bottomrule
\end{tabular}
\end{table}

\begin{table}[h]
\centering
\caption{Paragraph Level Evaluator Prompt}
\label{tab:paragraph_dce_prompt}
\begin{tabular}{p{13cm}}
\toprule
You are an evaluator. You will be given an article and a summary.  \\
The summary contains a summarized version of the article. \\
\\
\textbf{Criteria:} \\
Your task is to evaluate whether the summary is consistent with the article. You will evaluate it by going through the summary and checking against the following procedures:  \\ \\
* Understands all the aspects in the summary, who is doing what at when and where and what are the impact etc.  \\
* Compare if each aspect exists in the article  \\
* If it does, compare if the information in this summary is consistent with what is in the article  \\
* Compare if all the information in this summary can be directly inferred or entailed from what is in the article, including but not limited to who, what, when, where, etc.  \\
* It is OK that not all information from the article exists in this summary \\
 \\
\textbf{Given:}\\
\#\# Article \#\# \\
\{\textit{article}\} \\
\#\# Summary \#\# \\
\{\textit{summary}\} \\
\\
\textbf{Task} \\
Work in a step-by-step way to make sure we get the right answer. You will format the output in JSON as follows: \\
\{"reason": \{"sentence": "original sentence", "reason": "why this summary is or is not consistent with the article. You should start with \"This summary is consistent with the article\" or \"this summary is not consistent with the article\" "\},  "is\_consistent": true/false\} \\
\\
Here is the evaluation in JSON format:
\\
\bottomrule
\end{tabular}
\end{table}

\begin{table}[h]
\centering
\caption{Paragraph Level Reason Assisted Improver Prompt}
\label{tab:paragraph-improver-prompt}
\begin{tabular}{p{13cm}}
\toprule
You are a good writer. You will be given: \\
* an article \\
* a summary of the article \\
* a reason listing why the summary is or is not consistent with the article \\
    \\
Your job is to rewrite this summary: \\
* if the summary is consistent with the article, you can keep it as it is \\
* if the summary is not consistent with the article, you can rewrite it to make it consistent with the article based on the reasons given. \\
\\
\textbf{Article}\\
\{\textit{article}\}\\
\textbf{Summary} \\
\{\textit{summary}\} \\

\textbf{Reason} \\
\{\textit{reason}\} \\
\\
\textbf{Task}\\
Work in a step-by-step way to make sure we get the right answer. You will format the output in JSON as follows: \\
 \{"sentence": "original summary", "improved\_summary": "improved summary", "reason": "if it is improved, how it is improved. if not, say 'ALREADY CONSISTENT'"\}

\\
\bottomrule
\end{tabular}
\end{table}

\clearpage
\subsection{Divide-Conquer Evaluator Examples}

\begin{table}[h]
\centering
\begin{center}
\caption{QQP Evaluation Example}
\label{tab:qqp_example}
\begin{tabular}{m{3cm} | m{3cm} | m{2cm} | m{4cm}}
\toprule
\textbf{True Answer} & \textbf{Attempt Answer}  & \textbf{Is\_Consistent} & \textbf{Reasons}\\ \midrule
How do I read and find my YouTube comments? & How can I see all my YouTube comments? & True & The sentence is asking the same question as the true answer, just phrased differently. Both are asking about how to view one's own comments on YouTube \\ \midrule
What are the laws to change your status from a student visa to a green card in the US, how do they compare to the immigration laws in Canada? & What are the laws to change your status from a student visa to a green card in the US? How do they compare to the immigration laws in Japan? & False & \begin{itemize}[leftmargin=3mm]  
\item This sentence is consistent with the true answer as it asks about the laws to change status from a student visa to a green card in the US.
\item This sentence is not consistent with the true answer. The true answer asks for a comparison with the immigration laws in Canada, not Japan."   \end{itemize}  \\ \midrule

If I fire a bullet backward from an aircraft going faster than the bullet; will the bullet be going backward? & Do bullets travel faster than the speed of sound when shot from a gun? If not, is it possible? If they do, what gun and how much devastation occurs? & False & \begin{itemize}[leftmargin=3mm] \item This sentence is not consistent with the true answer because it is asking about the speed of bullets in relation to the speed of sound, not about the direction of a bullet when fired from a fast-moving aircraft. \item This sentence is not consistent with the true answer because it is asking about the possibility of bullets traveling faster than the speed of sound, not about the direction of a bullet when fired from a fast-moving aircraft. \item This sentence is not consistent with the true answer because it is asking about the type of gun and the amount of devastation, not about the direction of a bullet when fired from a fast-moving aircraft. \end{itemize} \\
\bottomrule
\end{tabular}
\end{center}
\end{table}

\begin{table}[h]
\centering
\begin{center}
\caption{PAWS Evaluation Example}
\label{tab:paws_example}
\begin{tabular}{m{3cm} | m{3cm} | m{2cm} | m{4cm}}
\toprule
\textbf{True Answer} & \textbf{Attempt Answer}  & \textbf{Is\_Consistent} & \textbf{Reasons}\\ \midrule
What were the major effects of the cambodia earthquake , and how do these effects compare to the Iquique earthquake in 1877 ? & What were the major effects of the Iquique earthquake , and how do these effects compare to the cambodia earthquake in 1877 ? & False & The sentence is not consistent with the true answer because it reverses the order of the earthquakes. The true answer asks about the effects of the Cambodia earthquake and compares it to the Iquique earthquake in 1877, while the attempt answer asks about the effects of the Iquique earthquake and compares it to the Cambodia earthquake in 1877.\\ \midrule
How do I start learning electronic music production ? Should I begin with books or videos ? I have logic pro 9 and I already made some songs but I would like to start from scratch ( as if I did not know anything ) . & How do I begin learning electronic music production ? Should I start with books or videos ? I have logic pro 9 and I already made some songs but I would like to start from scratch ( as if I did not know anything ) . & True & \begin{itemize}[leftmargin=3mm] \item This sentence is consistent with the true answer. The only difference is the use of the word 'begin' instead of 'start', but they mean the same thing. \item This sentence is consistent with the true answer. It is exactly the same. \item This sentence is consistent with the true answer. It is exactly the same. \end{itemize} \\
\bottomrule
\end{tabular}
\end{center}
\end{table}

\begin{table}[h]
\centering
\begin{center}
\caption{SummEval Evaluation Example}
\label{tab:summeval_example}
\begin{tabular}{m{2cm} | m{11cm}}
\toprule
 \textbf{Original Article}  & Paul Merson has restarted his row with Andros Townsend after the Tottenham midfielder was brought on with only seven minutes remaining in his team 's 0-0 draw with Burnley on Sunday . 'Just been watching the game , did you miss the coach ? \# RubberDub \# 7minutes , ' Merson put on Twitter . Merson initially angered Townsend for writing in his Sky Sports column that 'if Andros Townsend can get in ( the England team ) then it opens it up to anybody . ' Paul Merson had another dig at Andros Townsend after his appearance for Tottenham against Burnley Townsend was brought on in the 83rd minute for Tottenham as they drew 0-0 against Burnley Andros Townsend scores England 's equaliser in their 1-1 friendly draw with Italy in Turin on Tuesday night The former Arsenal man was proven wrong when Townsend hit a stunning equaliser for England against Italy and he duly admitted his mistake . 'It 's not as though I was watching hoping he would n't score for England , I 'm genuinely pleased for him and fair play to him ‚Äì it was a great goal , ' Merson said . 'It 's just a matter of opinion , and my opinion was that he got pulled off after half an hour at Manchester United in front of Roy Hodgson , so he should n't have been in the squad . 'When I 'm wrong , I hold my hands up . I do n't have a problem with doing that - I 'll always be the first to admit when I 'm wrong . ' Townsend hit back at Merson on Twitter after scoring for England against Italy Sky Sports pundit Merson ( centre ) criticised Townsend 's call-up to the England squad last week Townsend hit back at Merson after netting for England in Turin on Wednesday , saying 'Not bad for a player that should be 'nowhere near the squad ' ay @ PaulMerse ? ' Any bad feeling between the pair seemed to have passed but Merson was unable to resist having another dig at Townsend after Tottenham drew at Turf Moor . \\ 
 \midrule
 \textbf{Original Summary Divided into Bullet Points} & \begin{itemize} [leftmargin=3mm] 
  \item Paul merson was brought on with only seven minutes remaining in his team 's 0-0 draw with burnley.
   \item Andros townsend scored the tottenham midfielder in the 89th minute . 
    \item Paul merson had another dig at andros townsend after his appearance . 
     \item The midfielder had been brought on to the england squad last week .  
     \item Click here for all the latest arsenal news news .
     \end{itemize} \\\\ 
  \midrule
 \textbf{Reasons} & \begin{itemize} [leftmargin=3mm]  
  \item This sentence is not consistent with the article because it was Andros Townsend, not Paul Merson, who was brought on with only seven minutes remaining in the game.
 \item This sentence is not consistent with the article because it does not mention that Andros Townsend scored in the 89th minute. Also, the phrase 'scored the Tottenham midfielder' is not clear.   \item This sentence is consistent with the article. The article mentions that Paul Merson had another dig at Andros Townsend after his appearance.
\item This sentence is not consistent with the article because it does not mention that Andros Townsend was brought onto the England squad last week.
   \item This sentence is not consistent with the article because the article does not provide any latest Arsenal news.
  \end{itemize} \\
    \midrule
  \textbf{Is\_Consistent}  & False \\
\bottomrule
\end{tabular}
\end{center}
\end{table}

\begin{table}[h]
\centering
\begin{center}
\caption{QAGS-XSUM Evaluation Example}
\label{tab:qags_xsum_example}
\begin{tabular}{m{2cm} | m{11cm}}
\toprule
 \textbf{Original Article}  & London 's first history day will be held on the anniversary of big ben 's first day in operation . It will be first celebrated on 31 may in 2017 with celebrations and events run by historic england . The date was decided upon after a poll involving 1,000 londoners . It was closely followed by 5 september - the date of the great fire of london . The yougov questionnaire also declared the houses of parliament as the building that best sums up london . People voted for the queen as their favourite historic london hero for the moment she secretly joined the crowds to celebrate victory in europe day . The results of the poll were released to mark the launch of historic england 's `` keep it london '' campaign . People were asked to select a date to celebrate the capital 's history , their historic hero and the building that sums up london . Big ben 's first day in operation was 31 may 1859 . The campaign is intended to encourage londoners to notice , celebrate and speak up for the heritage of their city , historic england said . The public body has also launched a film entitled i am london , which celebrates the historic buildings and places that have borne witness to the capital 's history . Duncan wilson , chief executive of historic england , said : `` our heritage is a source of pride , a reminder of the city 's past , a foundation for its present and the building blocks for its future .  \\ 
 \midrule
 \textbf{Original Summary Divided into Bullet Points} & \begin{itemize} [leftmargin=3mm] 
  \item Big ben 's 150th anniversary has been chosen as the date to celebrate london 's history .
\end{itemize} \\\\ 
  \midrule
 \textbf{Reasons} & \begin{itemize} [leftmargin=3mm]  
  \item This sentence is not consistent with the article. The article states that London's first history day will be held on the anniversary of Big Ben's first day in operation, which is 31 May 1859. However, it does not mention that this date is the 150th anniversary of Big Ben.
  \end{itemize} \\
    \midrule
  \textbf{Is\_Consistent}  & False \\
\bottomrule
\end{tabular}
\end{center}
\end{table}

\begin{table}[h]
\centering
\begin{center}
\caption{QAGS-CNN Evaluation Example}
\label{tab:qags_cnn_example}
\begin{tabular}{m{2cm} | m{11cm}}
\toprule
 \textbf{Original Article}  & A southern iowa chiropractor accused of accepting sex as payment for his services and performing exorcisms on patients has surrendered his state license . The iowa board of chiropractic released a report wednesday detailing charges against charles manuel , of lamoni . Manuel signed an agreement last month admitting his misdeeds and pledging not to apply for reinstatement for at least 10 years . Patient satisfaction : a chiropractor in iowa has surrendered his license to practice and admitted to swapping services for sex and performing exorcisms on some patients . Nonetheless , he 's received outstanding evaluations for patient satisfaction on healthgrades . Com . The agreement requires that manuel prove any circumstances surrounding the revocation of his license no longer exist before he can resume practicing chiropractic in the state . Those circumstances included bartering sex for services with some patients . Manuel also recommended that patients stop taking medication he prescribed to them . A woman who answered a call to manuel 's home from the des moines register declined to comment on the case . A woman at his former practice said he had n't worked there for some time . A lamoni address listed on the yelp page for manuel 's practice appears to be a home on a residential street . While maneul has received just three patient survey responses on healthgrades . Com , those responses were quite positive . The disgraced chiropractor received a perfect five out of five stars in patient satisfaction . Strange practice : charles manuel , who admitted wrongdoing to the iowa board of chiropractic , listed his practice 's addresses on this residential street in the small agricultural town of lamoni in southern iowa .  \\ 
 \midrule
 \textbf{Original Summary Divided into Bullet Points} & \begin{itemize} [leftmargin=3mm] 
  \item A chiropractor in iowa has surrendered his license to practice and admitted to swapping services for sex and performing exorcisms on some patients. 
  \item Manuel also recommended that patients stop taking medication no longer exist before he can resume practicing chiropractic in the state .
  \item The disgraced chiropractor received a perfect five out of five stars in patient satisfaction .
\end{itemize} \\\\ 
  \midrule
 \textbf{Reasons} & \begin{itemize} [leftmargin=3mm]  
  \item This sentence is consistent with the article. The article states that Charles Manuel, a chiropractor in Iowa, surrendered his license and admitted to accepting sex as payment for his services and performing exorcisms on patients.
  \item This sentence is not consistent with the article. The article does mention that Manuel recommended patients stop taking medication he prescribed to them, but it does not state that this condition must be met before he can resume practicing.
  \item This sentence is consistent with the article. The article mentions that despite his misdeeds, Manuel received outstanding evaluations for patient satisfaction on healthgrades.com, including a perfect five out of five stars.']
  \end{itemize} \\
    \midrule
  \textbf{Is\_Consistent}  & False \\
\bottomrule
\end{tabular}
\end{center}
\end{table}

\clearpage

\subsection {Example of setting \(\alpha\) and \(\beta\) in equation \ref{eq:amc-final}}
\label{sec:alpha_beta}
\lstset{ %
  backgroundcolor=\color{white},   
  basicstyle=\footnotesize,        
  breakatwhitespace=false,         
  breaklines=true,                 
  captionpos=b,                    
  commentstyle=\color{commentsColor}\textit,    
  deletekeywords={...},            
  escapeinside={\%*}{*)},          
  extendedchars=true,              
  frame=tb,	                   	   
  keepspaces=true,                 
  keywordstyle=\color{keywordsColor}\bfseries,       
  language=json,                 
  otherkeywords={*,...},           
  numbers=left,                    
  numbersep=5pt,                   
  numberstyle=\tiny\color{commentsColor}, 
  rulecolor=\color{black},         
  showspaces=false,                
  showstringspaces=false,          
  showtabs=false,                  
  stepnumber=1,                    
  stringstyle=\color{stringColor}, 
  tabsize=2,	                   
  title=\lstname,                  
  columns=fixed                    
}

\(\alpha\) and \(\beta\) are added to ensure we are using a fully sentence level analysis. Below we give an example on how to use them to achieve this goal.

Suppose the output from Divide-Conquer Evaluator with customized prompt is: \\

\begin{lstlisting}[language=json,firstnumber=1]
{
  "is_consistent" : False, 
  "reasons" : [ 
    "The two paragraphs are not consistent.", 
    "This sentence is consistent." 
    "This sentence is not consistent." 
  ] 
}
\end{lstlisting}

After inspecting the {\em reasons} we notice that the first entry is not a sentence level analysis, but a paragraph level one. Thus when calculating scores we should remove the impact of this line.

Since in this case the first entry is negative, it will be given -1 score, thus we should set \(\alpha =1\) to mitigate this effect. Similarly, because we only have one entry that is not sentence level, we would set \(\beta = -1\).

Thus for this particular case, equation \ref{eq:amc-final} will be:

\begin{equation}
    \mathcal{Z} = \left(\sum_{i=1}^k z_i + 1 \right) / (k -1), \quad \hat{\mathcal{Z}} = (\mathcal{Z} + 1)/2, \quad \hat{\mathcal{Z}} \in [0,1]
\end{equation}

\clearpage

\subsection{What does LLM think of sentence level approach?}

To have some more fun, we asked different LLMs on what they think the best approach to evaluate consistency is, token level, paragraph level, or sentence by sentence level. We have a unanimous decision :)

\begin{table}[h]
\centering
\begin{center}
\caption{LLM's opinion on our sentence level approach}
\label{tab:llm_opinion}
\begin{tabular}{m{2cm} | m{11cm}}
\toprule
 \textbf{Input Prompt}  &  \textit{Given a reference, we have a generated paragraph from LLM. Our objective is to check whether the generated text is consistent to the reference or not. Can you you tell me, to achieve the best accuracy, would you choose to compare them token-by-token, direct check of paragraphs as a whole, or sentence by sentence check on the semantic meaning?}  \\  \\
 \midrule\textbf{GPT 3.5}  & To achieve the best accuracy in checking the consistency of generated text with a reference, \hl{it is recommended to compare them sentence by sentence on the semantic meaning}. This approach allows for a more nuanced analysis of the generated text and can identify inconsistencies that may not be apparent when comparing them token-by-token or as a whole paragraph. Additionally, comparing them sentence by sentence allows for a more efficient and targeted review of the generated text, as inconsistencies can be identified and addressed on a sentence-by-sentence basis.  \\  \\
 \midrule
 \textbf{GPT 4} &  \hl{To achieve the best accuracy, it would be most effective to check sentence by sentence on the semantic meaning.} This is because language can be complex and nuanced, and a direct token-by-token or whole paragraph comparison might miss subtle differences in meaning. By checking sentence by sentence, you can ensure that the generated text accurately reflects the reference in both content and context. \\ \\
  \midrule
 \textbf{PaLM2} &  To achieve the best accuracy in checking whether a generated paragraph from LLM is consistent with a reference, \hl{it would be best to compare them sentence by sentence on the semantic meaning.} This is because comparing token-by-token would be too time-consuming and would not take into account the overall meaning of the sentences. Direct check of paragraphs as a whole would also not be as effective, as it would not identify specific inconsistencies between the sentences. Comparing them sentence by sentence on the semantic meaning would allow for a more thorough and accurate check of the consistency between the generated paragraph and the reference. \\ \\
    \midrule
  \textbf{Claude 2}  &  \hl{Overall, I would recommend the sentence-by-sentence semantic comparison approach.} By checking the semantic consistency at a sentence level, it provides a good balance between catching inconsistencies in meaning while allowing for minor paraphrasing and word order changes. The paragraph-level comparison is too holistic while the token-by-token approach is too strict. Focusing on sentence-level meaning comparisons allows catching semantic inconsistencies with reasonably high accuracy. \\ \\
\bottomrule
\end{tabular}
\end{center}
\end{table}
 
\end{document}